\pdfoutput=1

\documentclass[11pt]{article}

\usepackage[final]{acl}

\usepackage{times}
\usepackage{latexsym}
\usepackage{tabularx}
\usepackage{amsmath}
\usepackage{enumitem}
\usepackage{subcaption}
\usepackage[T1]{fontenc}

\usepackage[utf8]{inputenc}

\usepackage{microtype}

\usepackage{inconsolata}

\usepackage{graphicx}
\usepackage{booktabs}
\usepackage{multirow}
\usepackage{array}
\usepackage{caption}
\usepackage{siunitx}

\NewDocumentCommand{\yi}
{ mO{} }{\textcolor{orange}{\textsuperscript{\textit{May}}\textsf{\textbf{\small[#1]}}}}
\NewDocumentCommand{\yumeng}
{ mO{} }{\textcolor{blue}{\textsuperscript{\textit{yumeng}}\textsf{\textbf{\small[#1]}}}}

\usepackage{tcolorbox}
\usepackage{amsmath}    
\usepackage{amssymb}   
\usepackage{bm}    

\definecolor{downcolor}{RGB}{255,102,102}
\definecolor{RoyalBlue}{RGB}{65,105,225}

\newcommand{\upcolor}[1]{\textcolor{RoyalBlue}{(+#1)}}
\newcommand{\downcolor}[1]{\textcolor{downcolor}{(-#1)}}

\title{Unveiling the Lack of LVLM Robustness to Fundamental Visual Variations: \\
Why and Path Forward
}

\author{Zhiyuan Fan\thanks{Equal contribution.} ~~~~Yumeng Wang$^*$ ~~~~Sandeep Polisetty ~~~~Yi R. (May) Fung \\
  Hong Kong University of Science and Technology \\
  \texttt{zhiyuan.fan@connect.ust.hk} ~~~\texttt{yrfung@cse.ust.hk} 
 \\}

\begin{document}
\maketitle

\begin{abstract}
Large Vision Language Models (LVLMs) excel in various vision-language tasks. Yet, their robustness to visual variations in position, scale, orientation, and context that objects in natural scenes inevitably exhibit due to changes in viewpoint and environment remains largely underexplored.
To bridge this gap, we introduce \textbf{V$^2$R-Bench}, a comprehensive benchmark framework for evaluating \textbf{V}isual \textbf{V}ariation \textbf{R}obustness of LVLMs, which encompasses automated evaluation dataset generation and principled metrics for thorough robustness assessment.
Through extensive evaluation on 21 LVLMs, we reveal a surprising vulnerability to visual variations, in which even advanced models that excel at complex vision-language tasks significantly underperform on simple tasks such as object recognition. Interestingly, these models exhibit a distinct visual position bias that contradicts theories of effective receptive fields, and demonstrate a human-like visual acuity threshold.
To identify the source of these vulnerabilities, we present a systematic framework for component-level analysis, featuring a novel visualization approach for aligned visual features. Results show that these vulnerabilities stem from error accumulation in the pipeline architecture and inadequate multimodal alignment. Complementary experiments with synthetic data further demonstrate that these limitations are fundamentally architectural deficiencies, scoring the need for architectural innovations in future LVLM designs.\footnote{Our code and data is available at \url{https://github.com/toward-agi/Visual-Variations-Robustness}.}
\end{abstract}

\begin{figure*}[t]
  \includegraphics[width=\textwidth]{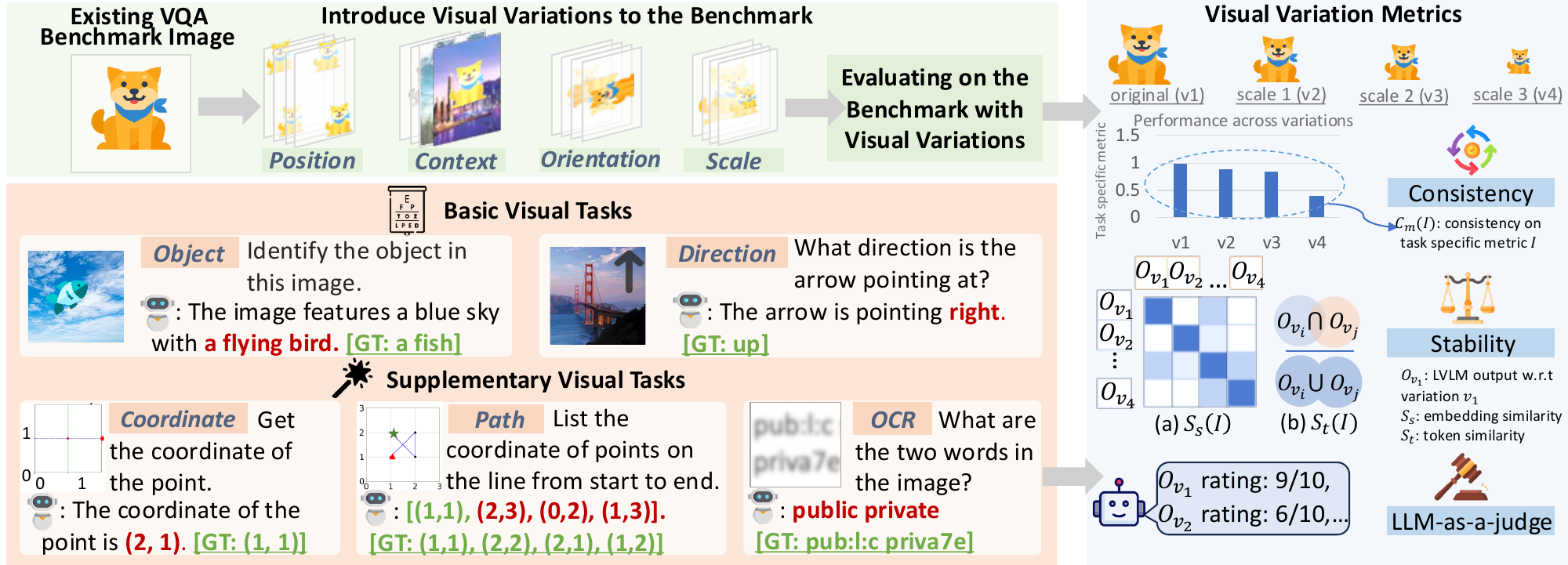}
  \caption{Our V$^2$R-Bench benchmark contains: 1) An automated data generation pipeline that adds visual variations to existing benchmarks; (2) Synthetic data generation for the fundamental visual tasks proposed ({\em e.g.,} object, direction, coordinate, path and OCR tasks).  The subfigure on the \textcolor{black}{\textit{right}} shows metrics for robustness testing, including performance consistency, semantic and token-level stability, and LLM-as-a-judge.} 
  \vspace{-9.6pt}
  \label{fig:pipeline}
\end{figure*}

\section{Introduction}
\vspace{-0.4em}
The rapid development of Large Vision Language Models (LVLMs) \citep{liu2023visualinstructiontuning,lu2024deepseekvlrealworldvisionlanguageunderstanding} has been driven by two key factors: innovations in model architectures and the availability of high-quality training data. These models have demonstrated impressive results in complex vision-language tasks, achieving human-level performance across various challenges~\citep{fei-etal-2024-multimodal}. To systematically evaluate such capabilities, numerous multimodal benchmarks have been developed to assess models' fundamental knowledge~\citep{fu2024blinkmultimodallargelanguage}, perceptual capability~\citep{wu2023vguidedvisualsearch,zhang2024mmerealworldmultimodalllmchallenge}, cognitive understanding~\citep{fu2024mmecomprehensiveevaluationbenchmark}, and reasoning skills~\citep{lu2024mathvistaevaluatingmathematicalreasoning,huang-etal-2024-lvlms} across various downstream applications.

While current benchmarks extensively evaluate models using images collected in specific scenarios, they overlook a more fundamental generalization capability:~\textbf{the robustness of LVLMs to visual variations}. In input images, objects naturally exhibit diverse variations: spatial positions shift with changes in camera angles and viewpoints; object scales vary depending on viewing distances; orientations deviate from standard poses through rotations and inversions; and objects appear in a range of visual semantic contexts. 
These visual variations raise concerns about LVLMs' robustness: \textit{whether models maintain consistent perception capabilities across all spatial positions in input images; what visual acuity threshold determines reliable performance; and how changes in orientation and visual context influence model behavior}. Despite the importance of such robustness, current research has primarily focused on model robustness to socio-cultural factors~\citep{ananthram2024perspectivediagnosingwesterncultural}, adversarial prompt attacks~\citep{wu2024on,liu2024pandoras}, or corrupted image inputs~\citep{liu2024are}, leaving the impact of natural visual variations largely unexplored. 

In this paper, we propose \textbf{V$^2$R-Bench}, a comprehensive benchmark framework to evaluate LVLM robustness against fundamental visual variations. Our framework consists of an automated data generation pipeline, and tailored evaluation metrics, which can readily extend to various VQA tasks. 
Through extensive evaluation on 21 LVLMs, we uncover surprising findings: despite their excellence in complex multimodal tasks, these models exhibit unexpected vulnerabilities to visual variations, leading to poor performance even in basic tasks such as object recognition. Specifically, we observe: (1) a counter-intuitive position bias where models achieve higher accuracy at image edges rather than the center; (2) a human-like visual acuity threshold where model reliability steadily decreases with object size, reaching and maintaining minimum performance below a critical scale threshold; (3) selective robustness to certain orientations while remaining fragile to others; and (4) a tendency to ground predictions on visual contextual inference rather than direct visual perception.
\vspace{-0.33em}

To identify the root cause of these vulnerabilities, we conduct a systematic analysis of LVLM components. A key innovation in our analysis is a novel visualization method that reconstructs language tokens from aligned visual features, offering insights into how models process and transmit visual information across modalities. Our investigation reveals that inadequate multimodal alignment is the primary bottleneck, as models fail to maintain stable visual representations across variations and struggle to effectively align visual semantics with the language model. To disentangle whether these limitations stem from architectural constraints or data deficiency, we conduct complementary experiments with synthetic training data. The results reveal that these vulnerabilities are fundamentally rooted in architectural design.
These findings underscore two critical directions for future LVLM development: stronger multimodal alignment mechanisms to maintain semantic consistency across variations, and unified architectural designs to mitigate error accumulation in current pipeline structures.

Our contributions are summarized as follows:
\begin{itemize}[leftmargin=*,noitemsep,topsep=0pt]
    \item We identify and formulate a novel problem: the robustness of LVLMs to visual variations, a fundamental yet overlooked capability essential for reliable vision-language reasoning.
    \item We propose V$^2$R-Bench, an evaluation framework with automated data generation and tailored metrics, uncovering significant vulnerabilities in current LVLMs through an extensive evaluation of 21 models.
    \item We develop a systematic component-level analysis with a novel visualization technique for multimodal alignment, revealing the root causes of these vulnerabilities and providing insights for future architectural improvements.
\end{itemize}

\section{Related Work}\label{appendix:related}
\vspace{-0.34em}

The rapid progress of LVLMs is driven by continuous advances in model architectures and the exponential growth of training datasets. On the architectural side, improvements in foundation language models~\citep{openai2024gpt4technicalreport,touvron2023llamaopenefficientfoundation,jiang2023mistral7b,qwen} and vision models~\citep{radford2021learningtransferablevisualmodels,oquab2024dinov2learningrobustvisual}, combined with increasingly sophisticated multimodal alignment modules~\citep{liu2023visualinstructiontuning,li2023blip2bootstrappinglanguageimagepretraining,alayrac2022flamingovisuallanguagemodel,zhu2023minigpt4enhancingvisionlanguageunderstanding,wang2024qwen2vlenhancingvisionlanguagemodels,zhou2024calibratedselfrewardingvisionlanguage}, have established the core capabilities of LVLMs in cross-modal understanding and reasoning. Building on these foundations, the introduction of high-quality datasets~\citep{chen2023sharegpt4vimprovinglargemultimodal,zhao2023svitscalingvisualinstruction,wang2023believepromptinggpt4vbetter,li2023m3itlargescaledatasetmultimodal} for different training stages (e.g., multimodal alignment, instruction tuning, preference alignment) has enabled LVLMs to achieve strong performance across diverse real-world scenarios. Despite effectively leveraging pretrained vision and language components, the pipeline architecture can accumulate errors across modules. No systematic investigation has yet been conducted to attribute model failures to specific components, limiting interpretability and understanding of modality alignment.

\paragraph{LVLM Evaluation and Benchmarking.}

Recent years have witnessed the emergence of numerous benchmarks for evaluating LVLM capabilities across cognitive and perceptual dimensions~\citep{liang2024hemmholisticevaluationmultimodal,fu2024blinkmultimodallargelanguage,fu2024mmecomprehensiveevaluationbenchmark,liu2024mmbenchmultimodalmodelallaround,yu2024mmvetevaluatinglargemultimodal,li2023seedbench2benchmarkingmultimodallarge,lee2024vhelmholisticevaluationvision,zhang2025vlm2benchcloserlookvlms}, assessing various aspects including reasoning skills, understanding abilities, and inherent knowledge. The scope of evaluation has further extended into specialized domains, with an increased emphasis on real-world scenario performance, as researchers develop dedicated benchmarks for embodied intelligence~\citep{wang2024mobileagentautonomousmultimodalmobile,yang2023gpt4toolsteachinglargelanguage,zhang2023appagentmultimodalagentssmartphone}, medical image analysis~\citep{xia2024carescomprehensivebenchmarktrustworthiness,chen2024gmaimmbenchcomprehensivemultimodalevaluation,hu2024omnimedvqanewlargescalecomprehensive}, chart understanding \cite{10787102}, and robotic control~\citep{chen2024rh20tpprimitivelevelroboticdataset,wang2023embodiedscanholisticmultimodal3d}. In parallel with capability assessment, researchers have begun investigating LVLM robustness from two critical perspectives: first, examining semantic biases in model responses, particularly those related to societal factors like gender and racial prejudices~\citep{wang2024vlbiasbenchcomprehensivebenchmarkevaluating,howard2024uncoveringbiaslargevisionlanguage,10.1145/3442188.3445932,wu2024macaroon}; and second, analyzing adversarial vulnerabilities via carefully crafted visual prompts for assessing model reliability under targeted attacks~\citep{liu2024are,luo2024jailbreakvbenchmarkassessingrobustness,zhang2024bavibenchevaluatingrobustnesslarge,wang2024vlbiasbenchcomprehensivebenchmarkevaluating,chen2023benchmarking}. Yet, the capacity to withstand basic visual variations, a fundamental aspect of robustness widely present in real-world deployment, remains unexplored in current research.

\section{V$^2$R-Bench Evaluation Framework}

\vspace{-5pt}

In this section, we propose a novel framework for automatically automatically generating diverse visual variations across task settings (\ref{sec:frame_generation}), along with a detailed description of the construction and characteristics of our evaluation dataset (\ref{sec:frame_dataset}) and our evaluation protocols (\ref{sec:frame_protocol}), as illustrated in Figure~\ref{fig:pipeline}.

\subsection{Automated Variations Generation}\label{sec:frame_generation}
The automated data generation pipeline incorporates four fundamental visual variations that are prevalent in real-world scenarios. \textbf{\textit{Position}} variations investigate whether LVLMs exhibit blind spots in their visual processing, where models may fail to accurately perceive visual information at certain spatial locations. Meanwhile, \textbf{\textit{scale}} variations examine the perceptual boundaries of LVLMs when processing objects at different scales, similar to clinical vision tests for humans. \textbf{\textit{Orientation}} variations challenge the ability of LVLMs to process objects at different rotational angles, which is crucial for real-world scenarios like robotic navigation where objects rarely appear in canonical orientations. \textbf{\textit{Context}} variations test whether model predictions remain consistent across diverse environmental settings, to reveal whether LVLMs perform genuine visual perception or rely primarily on contextual cues for inference.

Formally, given an image $I$, a set of transformed images along these dimensions is generated as:
\begin{equation}
    \mathcal{D} = \{T(I, v) | v \in \{P \times S \times R \times C\}\}
\end{equation}
\vspace{-0.05em}
\noindent where $P=\{1,...,W\} \times \{1,...,H\}$, $S=[s_{min}, s_{max}]$, $R=[0,2\pi]$, and $C \in \mathcal{B}$ represent the sets of position, scale, rotation, and context variations respectively. For each original question-image pair, this generation process produces $|P|\times|S|\times|R|\times|C|$ variants in total, enabling a holistic exploration of the entire variation space.

\subsection{Dataset Construction}
\label{sec:frame_dataset}

The proposed automated generation pipeline is implemented across two categories of tasks. 

The first category extends to existing multimodal benchmarks, focusing on scenarios where variations in position, scale, orientation, and context preserve ground-truth validity, ensuring that any performance changes reflect model robustness rather than ground-truth alteration. 

The second category examines fundamental visual capabilities through \textbf{\textit{object}} recognition and \textbf{\textit{direction}} detection tasks (i.e. basic visual tasks), with target objects and directional indicators being systematically transformed through our visual variations using image processing algorithms and inpainting diffusion models~\citep{10483967,lugmayr2022repaintinpaintingusingdenoising}. The \textbf{\textit{object}} dataset assesses a model's understanding of object positioning and directionality, which is a 90-class object classification task. For the \textbf{\textit{orientation}} dataset, the model performs an 8-class classification to detect an arrow's direction for evaluating its basic spatial reasoning.
 
The final evaluation datasets contain a total of 428K images. Each category serves a distinct yet complementary purpose in our evaluation: the basic tasks provide controlled, interpretable measures of fundamental capabilities, while the extended benchmarks assess robustness in more naturalistic settings. The detailed implementation of these generation algorithms is provided in Appendix~\ref{sec:experiment-set-up}.
\subsection{Evaluation Metrics}
\label{sec:frame_protocol}

In assessing LVLM robustness, we consider \textit{performance consistency}, which measures whether a model maintains its task-specific metrics across visual variations, by quantifying how these variations impact model performance:
\begin{equation}
C_m(I) = 1 - \sqrt{\frac{1}{N}\sum_{v \in \mathcal{V}} (M(I_v) - \overline{M})^2}
\end{equation}
where $M(I_v)$ denotes the task-specific metric on variation $I_v$, $\overline{M}$ represents the mean performance across all variations, and $\mathcal{V}$ is the set of variations. A larger $C_m$ indicates better consistency, with 1 representing that the model is robust enough to be unaffected by visual variations.

We also evaluate the \textit{output stability} of model generation, at both semantic and token levels: 
\vspace{-0.3em}
\begin{equation}
S_s(I) = \frac{1}{|\mathcal{V}|^2} \sum_{v_i \in \mathcal{V}} \sum_{v_j \in \mathcal{V}} sim(E(O_{v_i}), E(O_{v_j}))
\end{equation}

\vspace{-0.6em}
\begin{equation}
S_t(I) = \frac{1}{|\mathcal{V}|^2} \sum_{v_i \in \mathcal{V}} \sum_{v_j \in \mathcal{V}} \frac{|T(O_{v_i}) \cap T(O_{v_j})|}{|T(O_{v_i}) \cup T(O_{v_j})|}
\end{equation}

\noindent where $E(O)$ denotes the output embedding, $sim(\cdot,\cdot)$ computes the cosine similarity between embeddings, and $T(O)$ represents the set of tokens in output $O$. 

Furthermore, LLM-as-a-judge~\citep{zheng2023judgingllmasajudgemtbenchchatbot} is employed to emulate human assessment of LVLM-generated outputs under structured visual variations, providing an additional qualitative perspective on model robustness. 
The overall robustness is then determined by a weighted aggregation of these three evaluation dimensions.

\begin{table*}[t]
\small
\centering
\scalebox{0.818}{
\begin{tabular}{l|cccccccc|cccccccc}
\toprule
\multirow{3}{*}{\textbf{Model}} & \multicolumn{8}{c|}{\textbf{\textcolor{gray}{(Accuracy)}}} & \multicolumn{8}{c}{\textbf{\textcolor{gray}{(Robustness)}}} \\
 & \multicolumn{2}{c}{\textbf{Position}} & \multicolumn{2}{c}{\textbf{Orientation}} & \multicolumn{2}{c}{\textbf{Scale}} & \multicolumn{2}{c|}{\textbf{Context}} & \multicolumn{2}{c}{\textbf{Position}} & \multicolumn{2}{c}{\textbf{Orientation}} & \multicolumn{2}{c}{\textbf{Scale}} & \multicolumn{2}{c}{\textbf{Context}} \\
\cmidrule(lr){2-3} \cmidrule(lr){4-5} \cmidrule(lr){6-7} \cmidrule(lr){8-9} \cmidrule(lr){10-11} \cmidrule(lr){12-13} \cmidrule(lr){14-15} \cmidrule(lr){16-17}
 & \textit{Obj.} & \textit{Dir.} & \textit{Obj.} &\textit{Dir.} & \textit{Obj.} & \textit{Dir.} & \textit{Obj.} & \textit{Dir.}  & \textit{Obj.} & \textit{Dir.} & \textit{Obj.} &\textit{Dir.} & \textit{Obj.} & \textit{Dir.} & \textit{Obj.} & \textit{Dir.} \\
\midrule
Qwen-VL & 3.7 & 28.5 & 2.3 & 28.1 & 15.3 & 47.7 & 5.1 & 47.7 & 91.3 & 89.7 & 92.4 & 77.3 & 89.1 & 89.9 & 83.2 & 87.9 \\
Qwen2-VL-7B & 5.9 & 40.2 & 6.7 & 42.7 & 17.5& 50.9 & 6.7 & 42.7 & 94.5 & 91.0 & 95.5 & 66.4 & 86.2 & 92.3 & 93.7 & 89.0 \\[0.05em]
Molmo-7B-D & 26.0 & 62.6 & 26.2 & 64.6 & 27.8 & 63.2 & 26.2 & 64.6 & 91.1 & 91.0 & 95.5 & 66.4 & 86.2 & 92.3 & 93.7 & 89.0 \\
H2OVL-2B & 5.5 & 39.8 & 6.3 & 42.3 & 17.2 & 50.6 & 6.3 & 42.3 & 83.3 & 91.0 & 97.7 & 100 & 92.6 & 100 & 100 & 97.9 \\
H2OVL-800M & 1.3 & 20.1 & 1.1 & 19.9 & 17.9 & 33.2 & 17.3 & 30.8 & 79.3 & 89.0 & 92.1 & 74.5 & 83.2 & 92.6 & 89.5 & 90.1 \\ 
Phi3-Vision & 1.5 & 36.7 & 1.8 & 37.9 & 6.4 & 37.5 & 1.8 & 37.9 & 96.8 & 92.9 & 100 & 62.2 & 93.7 & 100 & 100 & 92.9\\
Phi3.5-Vision & 6.2 & 40.8 & 7.0 & 41.7 & 7.9 & 43.0 & 6.7 & 41.7 & 95.5 & 92.3 & 94.5 & 63.5 & 100 & 100 & 93.7 & 92.3\\
InternVL-Mono & 5.7 & 40.0 & 6.5 & 42.5 & 17.7 & 51.0 & 6.5 & 42.5 & 91.5 & 90.5 & 93.5 & 74.5 & 91.5 & 95.5 & 92.5 & 91.5 \\
InternVL-2 & 6.1 & 40.5 & 6.9 & 43.0 & 17.9 & 51.3 & 6.9 & 43.0 & 94.5 & 92.5 & 96.5 & 78.5 & 94.5 & 97.5 & 95.5 & 94.5 \\
InternVL-2.5 & 5.8 & 40.3 & 6.6 & 42.8 & 17.6 & 51.1 & 6.6 & 41.8 & 96.5 & 93.5 & 98.5 & 81.5 & 96.5 & 98.5 & 97.5 & 96.5 \\
LLaVA1.5 & 5.6 & 39.9 & 6.4 & 42.4 & 17.4 & 50.8 & 6.4 & 42.4 & 95.5 & 92.5 & 97.5 & 76.5 & 95.5 & 98.5 & 96.5 & 95.5 \\
LLaVA-Onevision-Qwen2 & 16.0 & 44.6 & 17.3 & 45.9 & 24.5 & 47.4 & 17.3 & 45.9 & 92.9 & 91.1 & 89.5 & 71.2 & 92.9 & 100 & 83.0 & 90.5 \\
LLaVA1.6-mistral & 1.7 & 19.0 & 1.9 & 19.4 & 4.4 & 21.6 & 1.9 & 19.4 & 96.8 & 93.7 & 100 & 78.8 & 96.8 & 96.8 & 100 & 96.8 \\
LLaVA1.6-vicuna & 1.8 & 19.2 & 2.0 & 19.6 & 4.5 & 21.8 & 2.0 & 19.6 & 96.5 & 93.5 & 98.5 & 77.5 & 96.5 & 97.5 & 98.5 & 96.5 \\
LLaMA3.2-11B-V & 5.7 & 40.0 & 6.5 & 42.5 & 17.7 & 51.0 & 6.5 & 42.5 & 92.5 & 91.3 & 94.5 & 75.2 & 92.5 & 96.5 & 93.0 & 92.5 \\
LLaMA3.2-90B-V & 6.3 & 41.2 & 7.1 & 43.7 & 18.2 & 51.7 & 7.1 & 43.7 & 97.5 & 94.5 & 98.6 & 82.5 & 97.5 & 98.4 & 98.5 & 97.5 \\
GLM4-V-9B & 14.1 & 27.9 & 9.3 & 11.7 & 38.1 & 25.7 & 1.2 & 23.3 & 95.3 & 92.5 & 97.1 & 79.6 & 95.9 & 97.0 & 96.1 & 95.5 \\
MiniCPM-V-2 & 5.7 & 39.2 & 1.3 & 39.3 & 15.6 & 41.9 & 6.3 & 42.8 & 93.2 & 91.2 & 95.2 & 76.2 & 93.7 & 96.3 & 94.8 & 93.1 \\
MiniCPM-V-2.5 & 9.2 & 44.8 & 9.0 & 48.3 & 21.0 & 55.5 & 9.0 & 46.3 & 95.3 & 92.2 & 97.3 & 78.2 & 95.7 & 97.2 & 96.8 & 94.8 \\
MiniCPM-V-2.6 & 5.9 & 40.5 & 6.7 & 43.0 & 17.8 & 51.3 & 6.7 & 43.0 & 96.2 & 93.3 & 98.2 & 79.3 & 96.8 & 98.3 & 97.7 & 95.9 \\[0.05em]
\midrule
GPT-4o & 31.5 & 78.2 & 29.8 & 77.2 & 31.9 & 69.5 & 26.7 & 79.9 & 98.3 & 96.2 & 100 & 85.3 & 98.7 & 100 & 100 & 98.2 \\
\bottomrule
\end{tabular}
}
\caption{Comprehensive evaluation of model performance ({\em i.e.,} accuracy and aggregated robustness) under different visual variations, for object and direction detection tasks) with metrics defined in Section~\ref{sec:frame_protocol}. } 
\label{tab:bias_tasks_basic}
\end{table*}
\section{Cross-Modality Diagnosis}

To understand the underlying mechanisms of LVLM vulnerabilities to visual variations, we propose a systematic analysis framework that examines contributions of each model components to robustness issues. Central to this framework is a novel visualization technique that provides straightforward insights into how visual features extracted by vision encoder are processed through the multimodal alignment module and aligned with language embedding space.

\subsection{Component-level Analysis}

Here we first formalize the general architecture and process pipeline of modern LVLMs to facilitate subsequent analysis. Given an input question $Q$ and an image $I$, the vision encoder $E_{v}$ first extract visual features from the input image:
\begin{equation}
v = E_{v}(I; \theta_{v}) \in \mathbb{R}^{N_v \times D_v}
\end{equation}
These visual features are then projected into the language embedding space through a multimodal alignment module $P$:
\begin{equation}
h = P(v; \theta_p) \in \mathbb{R}^{N_v \times D_h}
\end{equation}
Finally, the language model $M$ takes the aligned visual features in conjunction with question embeddings $E_{l}(Q) \in \mathbb{R}^{N_q \times D_h}$ as input to generate the response in an autoregressive manner:
\begin{equation}
R_t = M(h, E_{l}(Q), R_{<t}; \theta_m) \quad \text{for } t = 1, \dots, T
\end{equation}

As the initial input module of LVLMs, the \textbf{vision encoder} $E_v$ fundamentally determines the performance ceiling of the whole model, since the visual information contained in its extracted features represents the upper bound of visual content available to subsequent modules. Given the distinct pretraining paradigms of vision encoders (contrastive learning~\citep{radford2021learningtransferablevisualmodels} and self-supervised learning~\citep{oquab2024dinov2learningrobustvisual}), the feature quality assessment differs accordingly: for contrastive-trained encoders, the corresponding text encoder enables zero-shot analysis, while for supervised-trained ones, linear probing~\citep{alain2018understandingintermediatelayersusing} and clustering analysis serve to assess their extracted features.

The \textbf{multimodal projector} $P$ acts as a bridge between the visual feature and language embedding spaces, which raises two questions for its performance analysis: \textbf{(RQ1)} Do the projected features $h$ preserve the visual information contained in $v$, and \textbf{(RQ2)} Do the projected features $h$ align well with the embedding space of the language model $M$? To answer these questions, two analytical approaches are employed: (1) comparing the performance of pre-projection features $v$ and post-projection features $h$ on visual tasks to quantify potential visual information loss during multimodal alignment, and (2) measuring the discriminability between projected visual features $h$ and the language embeddings of their corresponding captions to assess the quality of modality alignment.

The \textbf{language model} $M$, as the final module for integrating aligned visual features with input questions and generating responses, is inherently affected by potential errors from upstream modules, which complicates the assessment of intrinsic language model robustness. Thus, an evaluation strategy is devised to bypass upstream modules by directly providing visual information as language tokens, which simulates ideal visual feature extraction and multimodal alignment where visual information is perfectly preserved without any loss or misalignment, enabling a controlled setting for analyzing inherent language model capabilities. 
We create text-based evaluation datasets mirroring LVLMs' visual tasks, using matrix-structured text to simulate ideally encoded visual scenes with diverse variations. Comparing performance on text-based versus visual tasks identifies if language models cause vulnerabilities to visual variations.


\subsection{Visual-Linguistic Feature Analysis}
\vspace{-0.25em}
In addition to quantitative analysis of multimodal alignment, our proposed framework also incorporates a novel visualization approach, which reconstructs language tokens from aligned visual features to provide interpretable evidence of the alignment process, an aspect previously unexplored. Specifically, given a single aligned visual feature $h \in \mathbb{R}^{1 \times D_h}$ and token embedding matrix $E \in \mathbb{R}^{|V| \times D_h}$ of the language model $M$, the feature $h$ can be decoded to a set of language tokens that approximate its semantic meaning:
\vspace{-0.55em}
\begin{equation}
t = \text{topk}\left(\text{softmax}\left(h E^\top\right)\right),
\end{equation}
where $|V|$ denotes the language model vocabulary size and $k$ controls the number of selected tokens.

Through these decoded language tokens, the semantic meaning captured by the aligned visual features becomes intuitively understandable, which helps address the following questions: \textbf{(RQ3)} how do language models interpret these aligned visual features? \textbf{(RQ4)} How robust is the semantics of aligned visual features to visual variations? These questions are investigated in detail in Section ~\ref{analysis_result}.

\begin{figure*}[t]
\centering
\vspace{-5mm}
\subfloat{\includegraphics[trim={4cm, 0cm 0cm 0cm},height=1.5in]{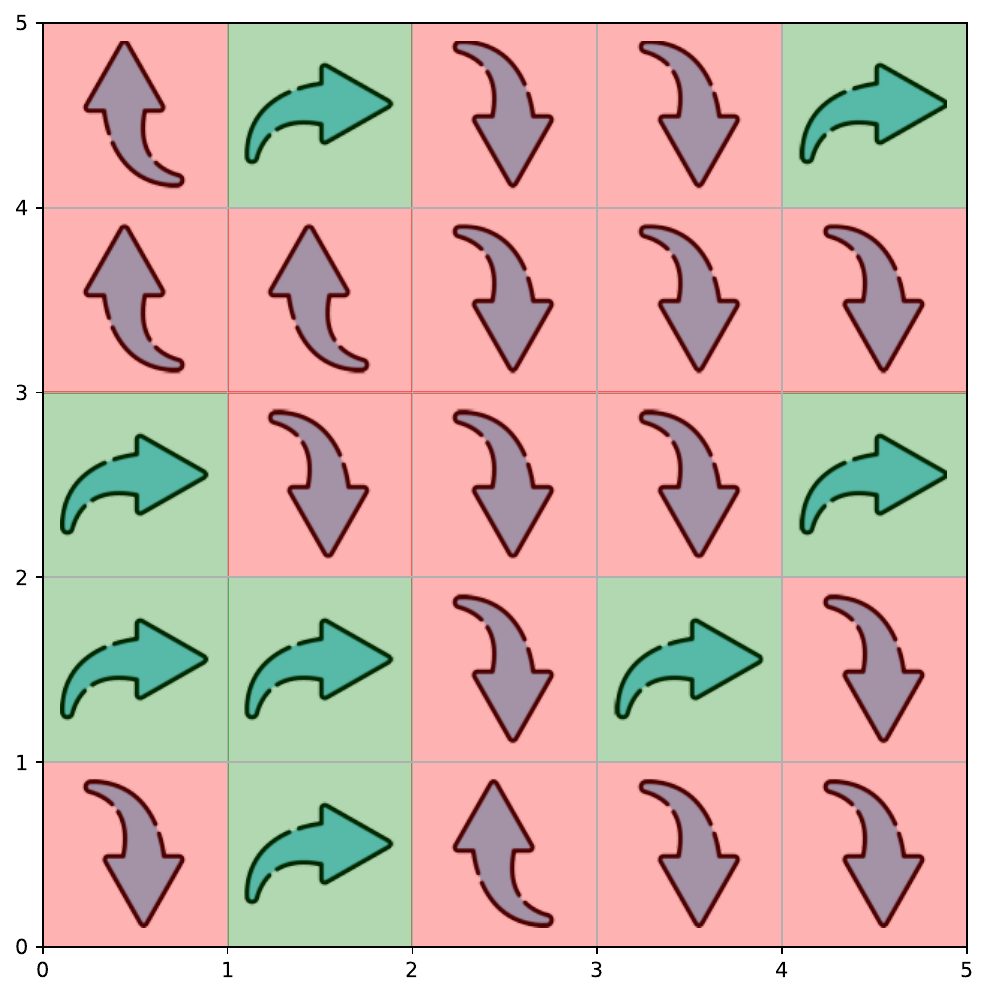}%
\label{fig:synthetic_p}}
\hfil
\subfloat{\includegraphics[trim={10cm, 0cm 2.7cm 0.16cm},height=1.6in]{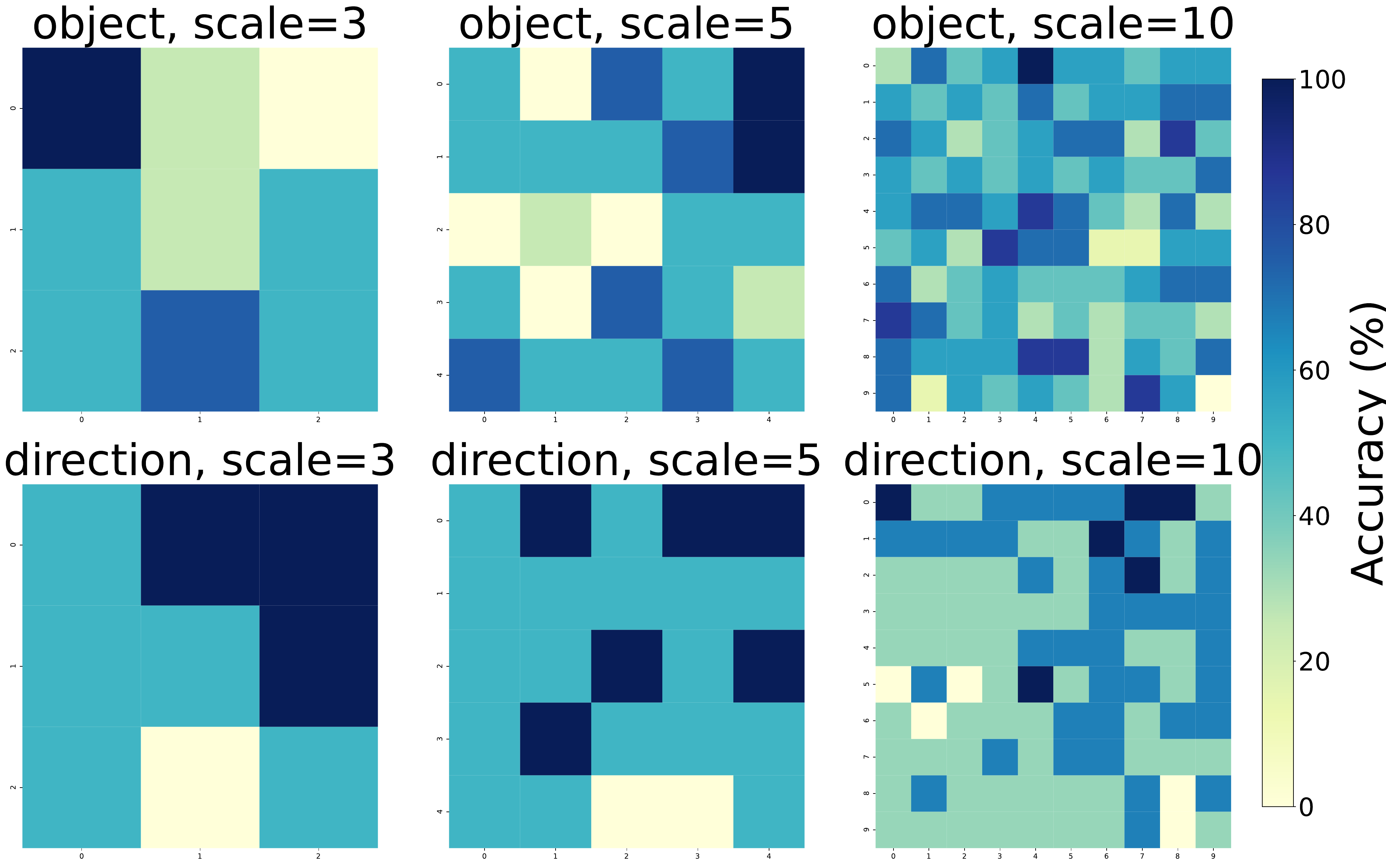}%
\label{fig:synthetic_l}}
\hfil
\subfloat{\includegraphics[trim={2.5cm, 0cm 4cm 0.16cm},height=1.5in]{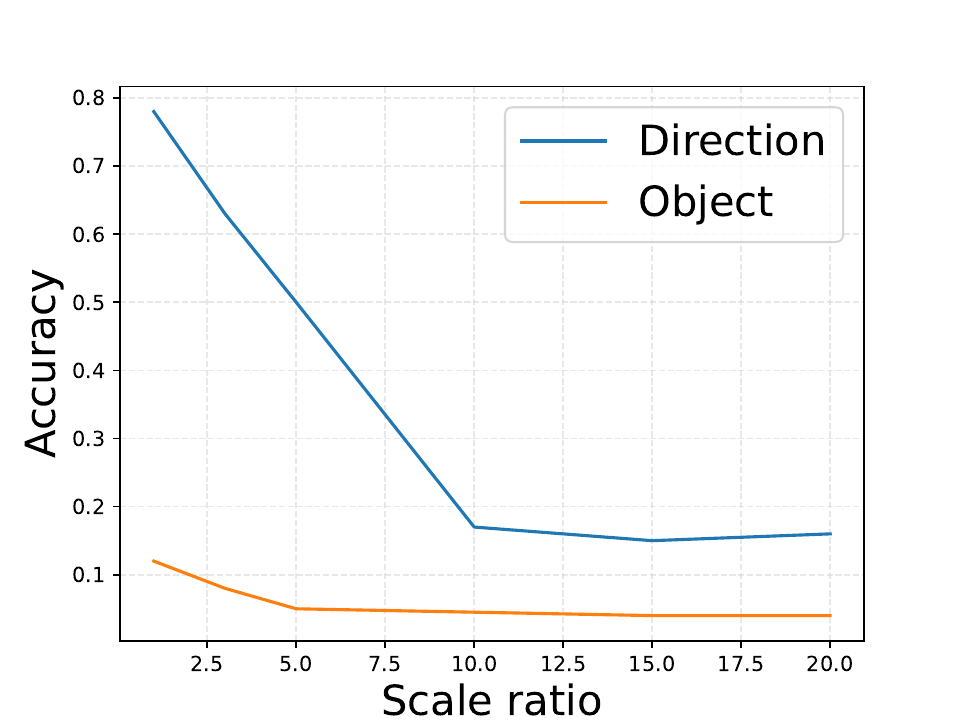}%
\label{fig:synthetic_s}}
\vspace{-0.5em}
\caption{\textit{(a)} Demonstration of position bias effects. \textit{~(b)} Accuracy heatmaps for object recognition and direction recognition, across object scales and position variations. \textit{~(c)} Model accuracy as a function of relative object scale.}
\label{fig:visual_variation_effects}
\vspace{-0.3em}
\end{figure*}
\section{Experiments}
\vspace{-0.3em}
Detailed experiment settings are provided in Appendix~\ref{app:exp}.

\subsection{Impact of Visual Variations}
The evaluation results across different LVLM architectures and model scales are presented in Table~\ref{tab:bias_tasks_basic}. Despite their impressive performance demonstrated on complex visual tasks, these models exhibit surprising vulnerability to simple visual variations, resulting in significantly degraded performance across basic visual tasks. Even proprietary models such as GPT-4o, Claude, and Gemini not only produce incorrect outputs, but also occasionally claim an inability to perceive the visual content entirely, as illustrated in Appendix~\ref{sec:examples}. Interestingly, despite claims that some carefully distilled smaller models outperform their larger counterparts on mainstream benchmarks~\cite{liu2024llavanext,Qwen-VL}, our analysis reveals that scaling laws still hold for robustness: within the same model architecture, larger models consistently demonstrate better stability across visual variations.

For an in-depth understanding of these vulnerabilities, this subsection focuses on LLaVA model, which provides full access to its training data, code, and parameters. \textbf{Position}: Figure~\ref{fig:visual_variation_effects}(b) shows that model performance varies dramatically across different positions. Contrary to the effective receptive field theory~\citep{luo2017understandingeffectivereceptivefield,raghu2022visiontransformerslikeconvolutional} which suggests vision models have better perception of central regions, LVLMs exhibit strong visual position bias with higher accuracy at peripheral regions (as shown in Table~\ref{tab:quant_pos_bias}), raising concerns about their fundamental visual processing mechanisms. Figure~\ref{fig:visual_variation_effects}(a) presents an intuitive example using right-pointing arrows as input, illustrating the significant effect of positional variations on prediction outcomes. \textbf{Scale}: Figure~\ref{fig:visual_variation_effects}(c) illustrates a sharp performance decline as object size decreases, 
\begin{table}[ht]
  \centering
  \resizebox{\linewidth}{!}{
    \begin{tabular}{lcccc}
      \toprule
      \multirow{2}{*}{\textbf{Model}} & \multicolumn{2}{c}{\textbf{Object}} & \multicolumn{2}{c}{\textbf{Direction}} \\
      \cmidrule(lr){2-3} \cmidrule(lr){4-5}
      & \textbf{Middle} & \textbf{Surrounding} & \textbf{Middle} & \textbf{Surrounding} \\
      \midrule
      LLaVA1.6 & 1.94 & 1.95 \upcolor{0.01} & 17.94 & 19.99 \upcolor{2.05} \\
      LLaVA1.5 & 1.83 & 1.89 \upcolor{0.06} & 15.33 & 18.02 \upcolor{2.69} \\
      Vila & 1.87 & 1.93 \upcolor{0.06} & 16.85 & 17.83 \upcolor{0.98} \\
      Ovis & 2.07 & 2.11 \upcolor{0.04} & 23.77 & 31.08 \upcolor{7.31} \\
      InternVL & 2.01 & 2.79 \upcolor{0.78} & 21.58 & 30.26 \upcolor{8.68} \\
      Qwen2-VL & 5.78 & 7.02 \upcolor{1.24} & 39.36 & 44.08 \upcolor{4.72} \\
      LLaMA3.2-11B-V & 3.72 & 6.58 \upcolor{2.86} & 27.19 & 35.88 \upcolor{8.69} \\
      \bottomrule
    \end{tabular}
  }
  \caption{Quantitative analysis of position bias on the proposed basic visual tasks: object detection and direction recognition. Experimental results show that these models consistently demonstrate better perception of surrounding regions compared to the central areas. 
  }
  \vspace{-1em}
  \label{tab:quant_pos_bias}
\end{table}
stabilizing when the object occupies 1/100 of the image area (equivalent to 1/10 of both width and height). This reveals a potential visual acuity threshold in LVLMs, similar to human vision, that defines a critical boundary below which model outputs become unreliable, serving as a key indicator for deploying LVLMs in fine-grained visual perception tasks. \textbf{Orientation}: performance degradation is observed across different orientations, with models exhibiting distinct directional biases: some orientations show robustness, while others lead to significant failures. Interestingly, models like Fuyu and BLIP demonstrate a pronounced predictive tendency, being heavily influenced by the left orientation. \textbf{Context}: model predictions vary with different contextual arrangements, correlating with context content, which raises questions about whether the model truly perceives the target objects or infers them from contextual cues.

To further investigate the root causes of these vulnerabilities, we design complementary tasks focusing on fundamental perceptual capabilities. To examine whether position and direction vulnerabilities stem from limitations in spatial and directional perception, we introduce two specialized tasks: coordinate identification, where models directly output position coordinates to test spatial understanding, and path tracking, where models sequentially output coordinates following directional lines to examine continuous directional perception. To verify whether models achieve genuine recognition or merely rely on contextual inference, we develop a modified OCR task where certain letters in fluent text are intentionally replaced with incorrect characters and then blurred at different levels, testing whether models faithfully report the visual content. 

\begin{figure}[h]
  \vspace{-0.7em}
  \center
  \includegraphics[trim={0cm, 0.5cm 0cm 0cm},width=0.9\columnwidth]{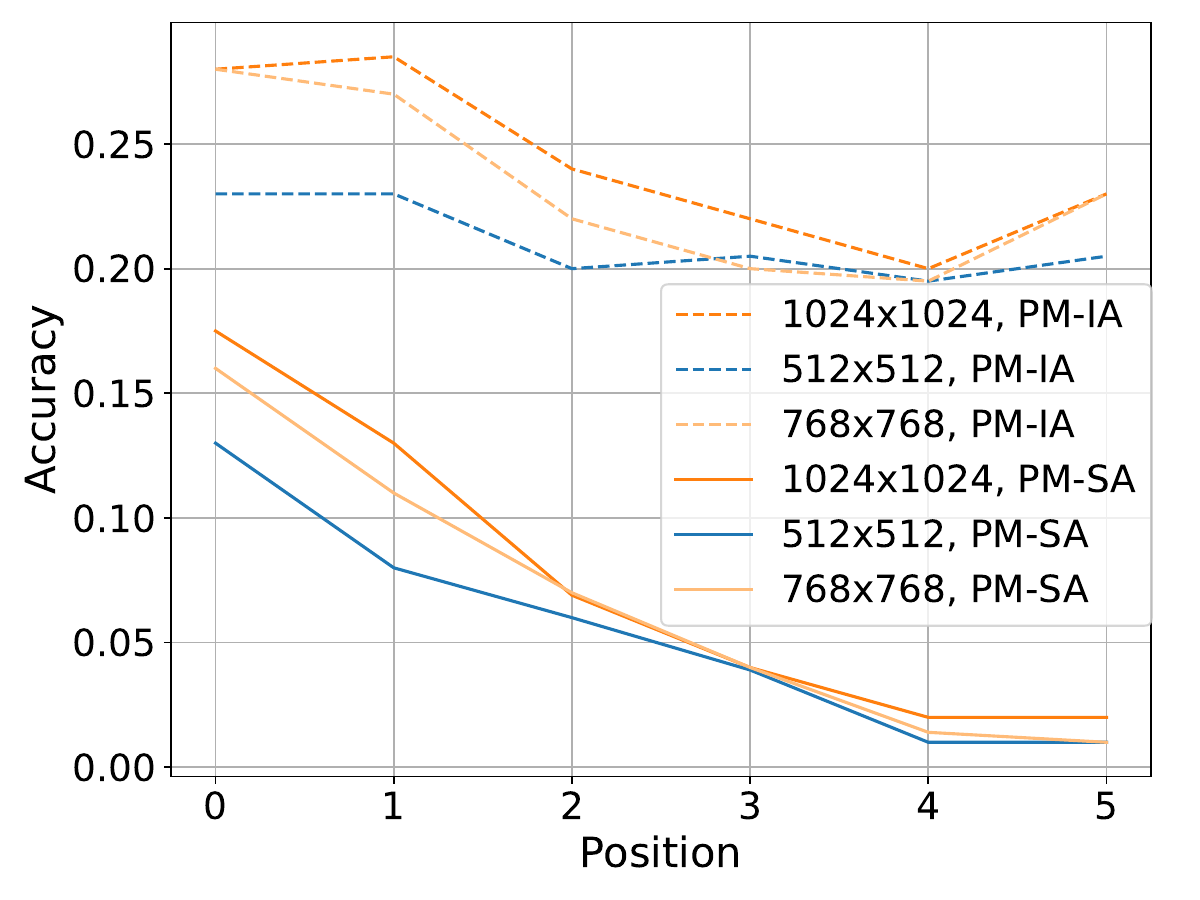}
  \caption{Performance evaluation on a 6-point path tracing task, where accuracy indicates coordinate prediction precision at each sequential point.} 
  \label{fig:path_acc_vs_pos}
  \vspace{-0.6em}
\end{figure}

As shown in Table \ref{table:metrics_eval_upper}, all models exhibit poor performance in coordinate recognition and path tracing, indicating these vulnerabilities fundamentally originate from the inability to accurately perceive spatial properties. Figure~\ref{fig:path_acc_vs_pos} reveals peak accuracy at initial points followed by monotonically decreasing performance, with marginal recovery at terminal nodes. This indicates not only weak isolated direction recognition but also inadequate performance in visual-following reasoning tasks, suggesting that spatial reasoning capabilities likely derive more from commonsense world knowledge rather than genuine visual understanding. Table~\ref{tab:ocr} results further demonstrate that LVLMs tend to overconfidently output contextually inferred conclusions rather than objectively depicted visual content, while LLMs achieve better accuracy under the same task conditions. This contrast proves that the observed limitations stem not from denoising characteristics~\citep{devlin-etal-2019-bert} inherent to the transformer architecture itself, but rather from interference introduced by visual modality integration.

\begin{table}[h]
\centering
\vspace{-0.1em}
\resizebox{\linewidth}{!}{
\begin{tabular}{lcccc}
\toprule
\textbf{Model} & EMA & PM-IA & PM-SA & PA \\
\midrule
Qwen2-VL & 6.0 & 23.9 & 7.9  & 48.7  \\
Molmo-7B-D & 0.0  & 16.0   & 5.3   & 19.9   \\
Phi3-Vison & 6.2  & 44.8  & 18.8  & 70.7  \\
Phi3.5-Vison  & 0.5 & 9.3  & 3.0  & 12.3 \\
LLaVA-Onevision-Qwen2 & 0.5  & 7.4 & 2.1 & 7.7  \\
LLaVA1.6 & 0.0 & 1.7 &  0.5 & 1.3 \\
\bottomrule
\end{tabular}
}
\caption{Path and coordinate task accuracy in \%, of 6 selected models across metrics as defined in Appendix~\ref{sec:path-dataset}.} 
\label{table:metrics_eval_upper}
\vspace{-0.15em}
\end{table}

\subsection{Component Analysis} \label{analysis_result}
\begin{figure}[t]
  \includegraphics[width=\columnwidth]{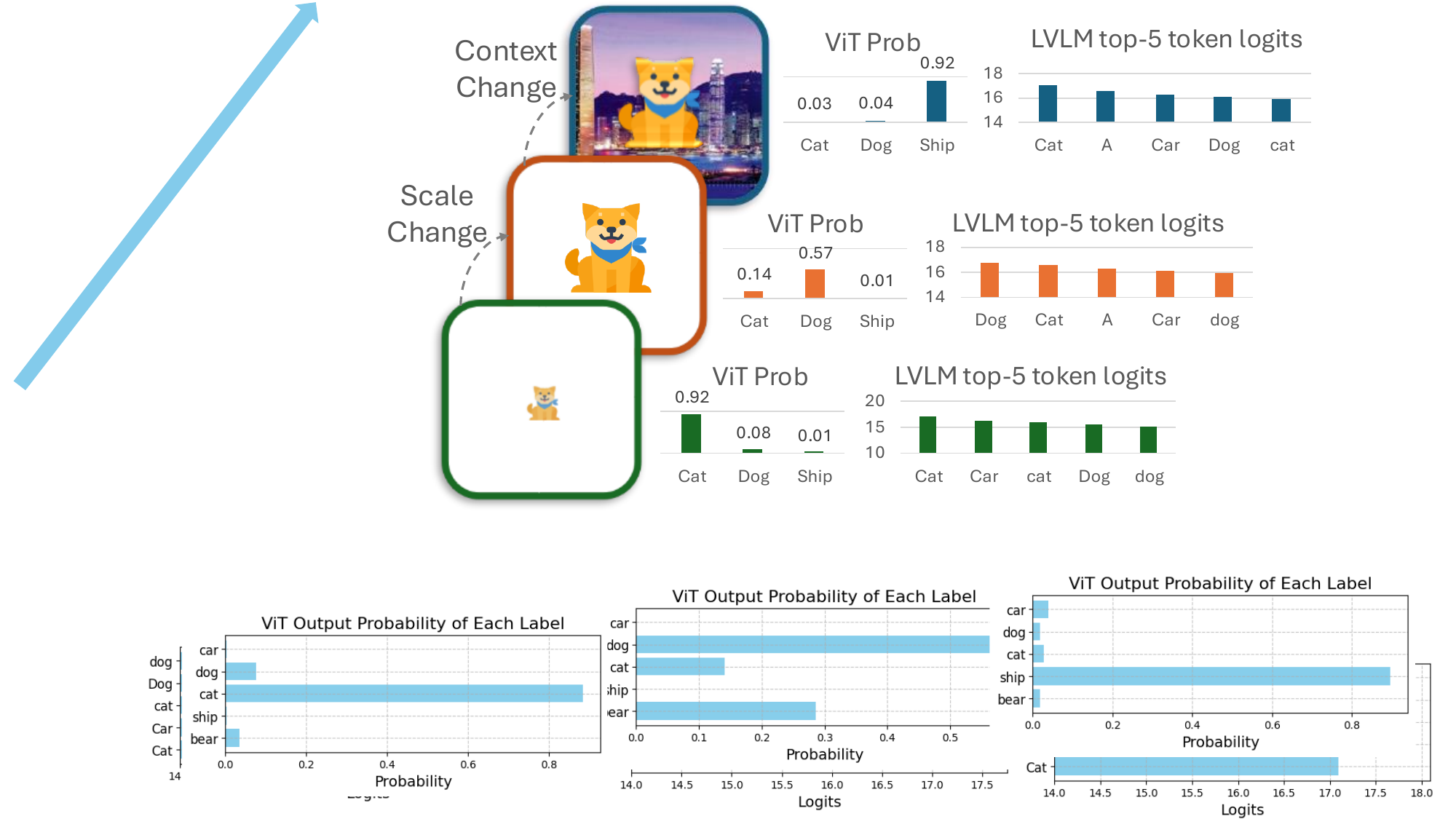}
  \caption{Changes in vision encoder classification probabilities and LVLM token predictions under different visual variations (context and scale). For each variation, we show the ViT's top-3 class probabilities (\textit{left}) and LVLM's top-5 token logits (\textit{right}), demonstrating how semantic interpretations shift across visual variations.}
  \label{fig:token_prob_change}
  \vspace{-0.8em}
\end{figure}

\paragraph{Vision Encoder.}
Figure~\ref{fig:token_prob_change} reveals that changes in the prediction label probabilities of the vision encoder align with shifts in the next token logits of the LVLM when scale and context variations are introduced. This behavioral consistency suggests that LVLM inherits vulnerability to visual variations from its vision encoder component.

\paragraph{Multimodal Projector.} The following analyses address the four research questions posed above.

\noindent\textbf{(RQ1) Visual Information Loss.}
Linear probing results for pre-projection and post-projection features are presented in Table~\ref{table:rebuttal_vit_mm_proj}. The significant performance degradation after multimodal projection ({\em i.e.,} the MM-Projector) suggests irrecoverable semantic information loss during the modality alignment process, contributing to LVLM's vulnerability across visual variations. We further compare the fine-tuned vision encoder in LLaVA 1.6 with the original one. While its vision encoder is fine-tuned to adapt image-text features for generative tasks, its linear probing performance underperforms the original vision encoder, revealing a trade-off: task-specific adaptation improves multimodal coherence but erodes the vision encoder's innate spatial representational fidelity. This aligns with LLaVA 1.5's limitations, where only the lightweight projector can be fine-tuned and struggles to mitigate information loss from the frozen CLIP encoder. Our results underscore that while recent advancements in alignment neural networks alleviate alignment bottlenecks, fundamental architectural constraints such as patch-based tokenization and positional bias persist as critical vulnerabilities, necessitating unified approaches to relax the conflicts between visual grounding and multimodal alignment.

\begin{table}[h]
\vspace{0.5em}
\centering
\resizebox{\linewidth}{!}{
\begin{tabular}{lllll}
\toprule
\multirow{2}{*}{\textbf{Model}} & \multicolumn{2}{c}{\textbf{ViT-LVLM}} & \multicolumn{2}{c}{\textbf{MMProj-LVLM}} \\
\cmidrule(lr){2-3} \cmidrule(lr){4-5}  
 &  Obj. & Dir. & Obj. & Dir.   \\
\midrule
LLaVA1.5 & 44.2 & 82.6 & 3.2~~~\downcolor{41.0} & 11.7 \downcolor{70.9} \\
LLaVA1.6 & 40.6 & 91.2 & 14.1 \downcolor{26.5} & 20.7 \downcolor{70.5} \\
Vila & 21.9 & 96.5 & 9.7~~~\downcolor{12.2} & 13.2 \downcolor{83.3} \\
InternVL & 22.1 & 96.4 & 13.2 \downcolor{8.9} & 33.9 \downcolor{62.5} \\
Ovis & 21.9 & 96.5 & 13.3 \downcolor{8.6} & 31.8 \downcolor{64.7} \\
Qwen2-VL & 44.6 & 100 & 65.9 \upcolor{21.3} & 100~~(0) \\
LLaMA3.2-11B-V & 22.2 & 100 & 11.9 \downcolor{10.3} & 98.6 \downcolor{1.4} \\
\bottomrule
\end{tabular}
}

\caption{Linear probing accuracy (\%) of vision language models on \textit{Object} and \textit{Direction} tasks reveals the impact of vision encoders/multimodal projectors in LVLMs. Notably, Qwen2-VL, which employs a native vision encoder trained directly with the language model rather than pretraining on image modality first and then aligning to language modality, shows no performance loss and even improvement after modality projection, suggesting the potential of unified architectures.}
\vspace{-1.15em}
\label{table:rebuttal_vit_mm_proj}
\end{table}

\begin{figure*}[ht]
  \small
  \centering
  \begin{minipage}[t]{0.3\textwidth}
    \centering
    \includegraphics[trim={0.8cm, 0.8cm, 0.5cm, 0.6cm}, clip, width=\linewidth]{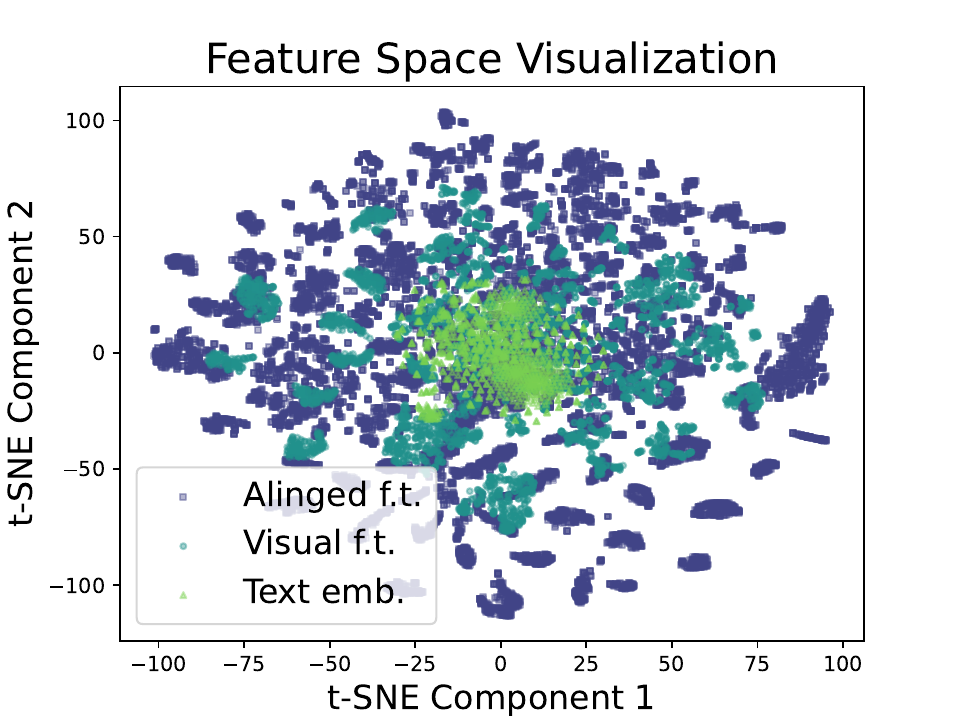}
    \label{fig:aligned_image_text_features}
  \end{minipage}
  \begin{minipage}[t]{0.32\textwidth}
    \centering
    \includegraphics[trim={0.1cm, 3.25cm, 0cm, 0cm}, clip, width=\linewidth]{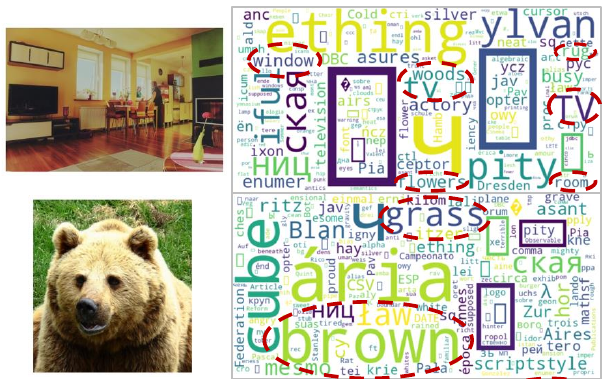}
    \label{fig:wordcloud}
  \end{minipage}
  \begin{minipage}[t]{0.36\textwidth}
    \centering
    \includegraphics[width=\linewidth]{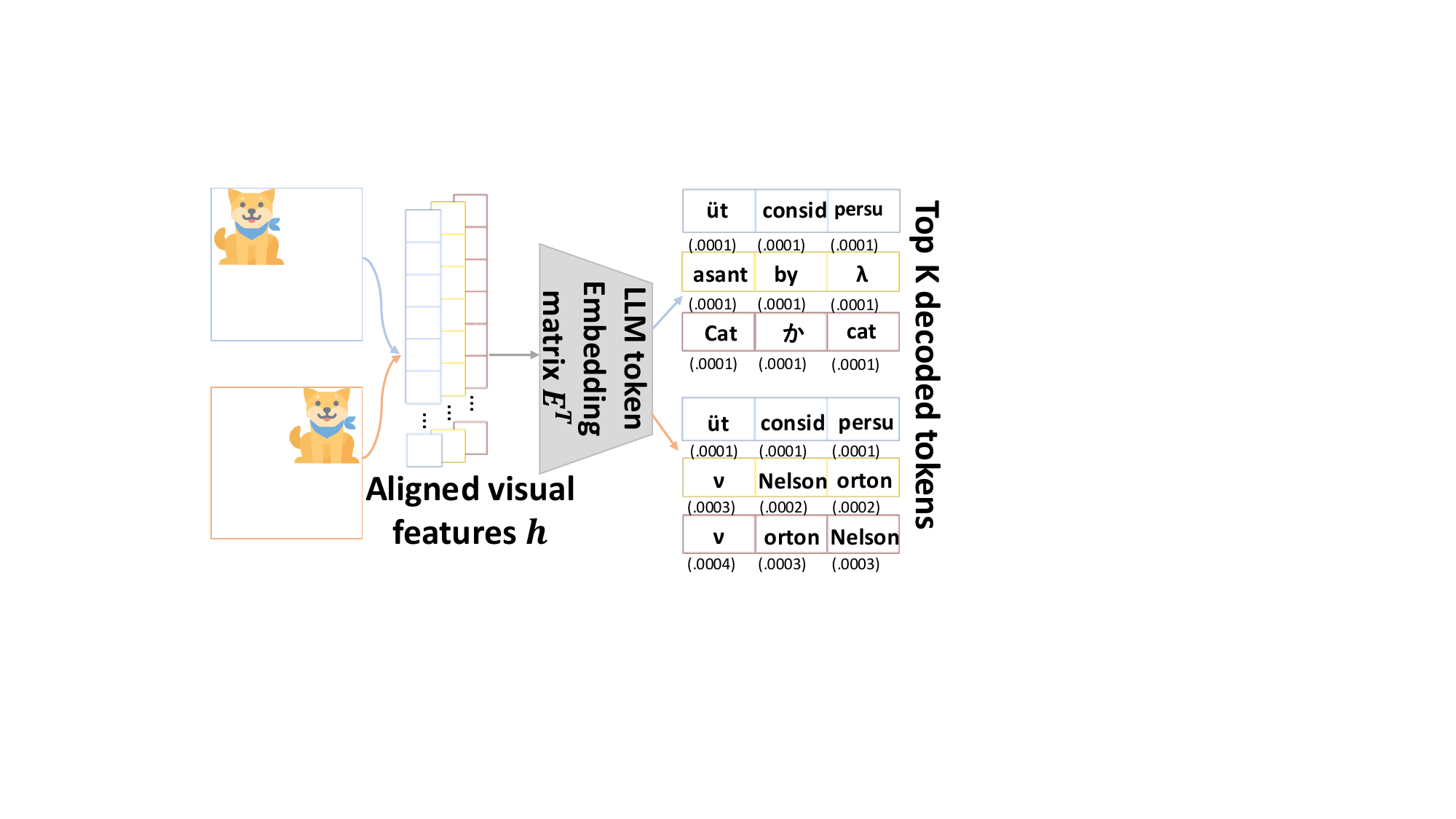}
    \label{fig:visual-linguistic}
  \end{minipage}
  \vspace{-0.9em}
  \caption{Comparison of different feature representation visualizations. (a) Visualization of visual features, aligned features and text embeddings. (b) Word cloud of decoded features. (c) Visual-linguistic token alignment.}
  \vspace{-1.2em}
  \label{fig:fig_page_7}
\end{figure*}

\noindent\textbf{(RQ2) Inadequate Multimodal Alignment.}
Figure \ref{fig:fig_page_7}(a) illustrates the spatial distribution of visual features, aligned features, and language embeddings. We observe that the distribution of the aligned features is similar to that of image features, meaning that the multimodal process preserves the visual semantics for some extent. Ideally, the aligned features should closely resemble the language embeddings. However, our observation reveals a significant disparity between the aligned feature space and the language embedding space, suggesting a lack of adequate modality alignment. This finding underscores the critical role of the multimodal projector, which is a key factor contributing to the LVLM's vulnerability to visual variations. To explore the image feature representation with regard to visual variations, Figure~\ref{fig:cluster_by_position_and_direction} in the appendix presents the clustering analysis of image features of the same object at different directions and positions. These variations introduce substantial alterations in image features, highlighting the inherent challenges faced by robust image encoding techniques. These fluctuations can lead to inconsistencies in feature extraction and representation across different instances of the same object or scene. The limited robustness in current encoding methodologies may struggle to effectively capture and encode these variations while maintaining the semantic integrity of the features, potentially compromising the accuracy and reliability of downstream tasks.
 Addressing this issue requires the development of more resilient encoding strategies that can adapt to diverse visual transformations, enhancing the overall robustness and generalizability of image feature representations in complex visual tasks.


\noindent\textbf{(RQ3) Interpretation of Aligned Features.}
Figure~\ref{fig:fig_page_7}(b) demonstrates the results of decoding aligned visual features into language tokens. Due to the inherent differences in attention patterns between vision encoder (bidirectional) and language model (autoregressive), these decoded results do not form coherent natural language, with only a subset of tokens being semantically relevant to the image content. These aligned visual features reside outside the discrete language embedding space, echoing back to the visualization shown in Figure~\ref{fig:fig_page_7}(a). This can be interpreted as visually conditioned soft prompts~\citep{lester2021powerscaleparameterefficientprompt,gu-etal-2022-ppt,liu-etal-2022-p}, providing implicit cues that contain image-related information to guide model responses to image-related queries, see Appendix~\ref{app:soft_prompt}. However, prior work~\citep{bailey2023soft} has revealed the limitations of such soft prompt approaches, including their instability and bugs due to lack of interpretability and the discrepancy from language model embedding space, ultimately leading to the vulnerability to visual variations.

\noindent\textbf{(RQ4) Visual Semantics Vulnerability.}
Based on the results for RQ3, significant changes in decoded language tokens are observed across visual variations, indicating poor semantic robustness, as shown in Figure~\ref{fig:fig_page_7}(c). These inconsistent soft prompt prefixes consequently lead to substantial fluctuations in model outputs, further compromising model robustness.
\vspace{-0.5em}
\paragraph{Language Model.}
It has come to the community's attention that positional biases in natural language widely exist in language models \cite{wang2023chatgptgoodnlgevaluator,jung-etal-2019-earlier,zheng2023judgingllmasajudgemtbenchchatbot,koo2024benchmarkingcognitivebiaseslarge}, where models tend to favor information appearing earlier in the text. As suggested in Table~\ref{table:llm_component}, we observe that language models exhibit position bias in matrix-structured text representations of visual information. Although object and background information are explicitly represented as language tokens, model predictions are still influenced by context, albeit to a lesser extent than LVLMs, suggesting that the vulnerability primarily stems from the upstream vision encoder and multimodal alignment module, where the aligned visual features fed into the language model are already heavily influenced by visual context. The low accuracy in the coordinate task demonstrates that, similar to LVLMs, language models also lack precise positional awareness of objects. These limitations are rooted in the autoregressive nature of language models, which tends to prioritize sequential dependencies over structural relationships. 
\begin{table}[h]
\vspace{0.22em}
\centering
\resizebox{\linewidth}{!}{
\begin{tabular}{lcccccc}
\toprule
\multirow{2}{*}{\textbf{Model}} & \multicolumn{2}{c}{\textbf{Number}} & \multicolumn{2}{c}{\textbf{Coordinate}} & \multicolumn{2}{c}{\textbf{Object}}  \\
\cmidrule(lr){2-3} \cmidrule(lr){4-5} \cmidrule(lr){6-7} 
 & w/o BG & w/ BG & w/o BG & w/ BG & w/o BG & w/ BG \\
\midrule
LLaMA3-8B & 6.88 & 10.12 & 0.32 & 0.24  & 99.96 & 84.64  \\
LLaMA3.1-8B & 17.42 & 13.06 & 0.80 & 0.27  & 97.54 & 91.89  \\
Mistral-v0.2 & 29.40 & 7.76 & 0.80 & 0.00  & 88.32 & 85.28  \\
Qwen-2 & 32.88 & 13.64 & 0.88 & 0.84 & 99.52 & 97.84 \\
Qwen-2.5 & 10.19 & 8.11 & 5.45 & 1.99 & 99.44 & 97.19 \\
Deepseek-7B & 18.26 & 16.33 & 1.06 & 0.93 & 85.43 & 41.71 \\
\bottomrule
\end{tabular}
}
\caption{Accuracy of LVLM's language model backbones on text-based tasks: comparing scenarios with tokenized background represented as asterisks (w/ BG) versus random words (w/o BG).}
\label{table:llm_component}
\vspace{-0.9em}
\end{table}

\section{Mitigating Robustness Issues}
\vspace{-0.45em}
To determine whether the robustness issues stem from architectural limitations or insufficient training data, we explore improvements through two complementary approaches: First, we conduct controlled experiments on a subset of our test data while maintaining a held-out test set, directly probing the architectural capacity for robust visual understanding. Second, we utilize a general visual instruction tuning dataset injected with visual variations to analyze whether a more diverse training distribution can enhance model robustness. 

\begin{figure}[!t]
\vspace{-0.1em}
    \includegraphics[trim={0cm, 0.6cm 0cm 0cm},width=\linewidth]{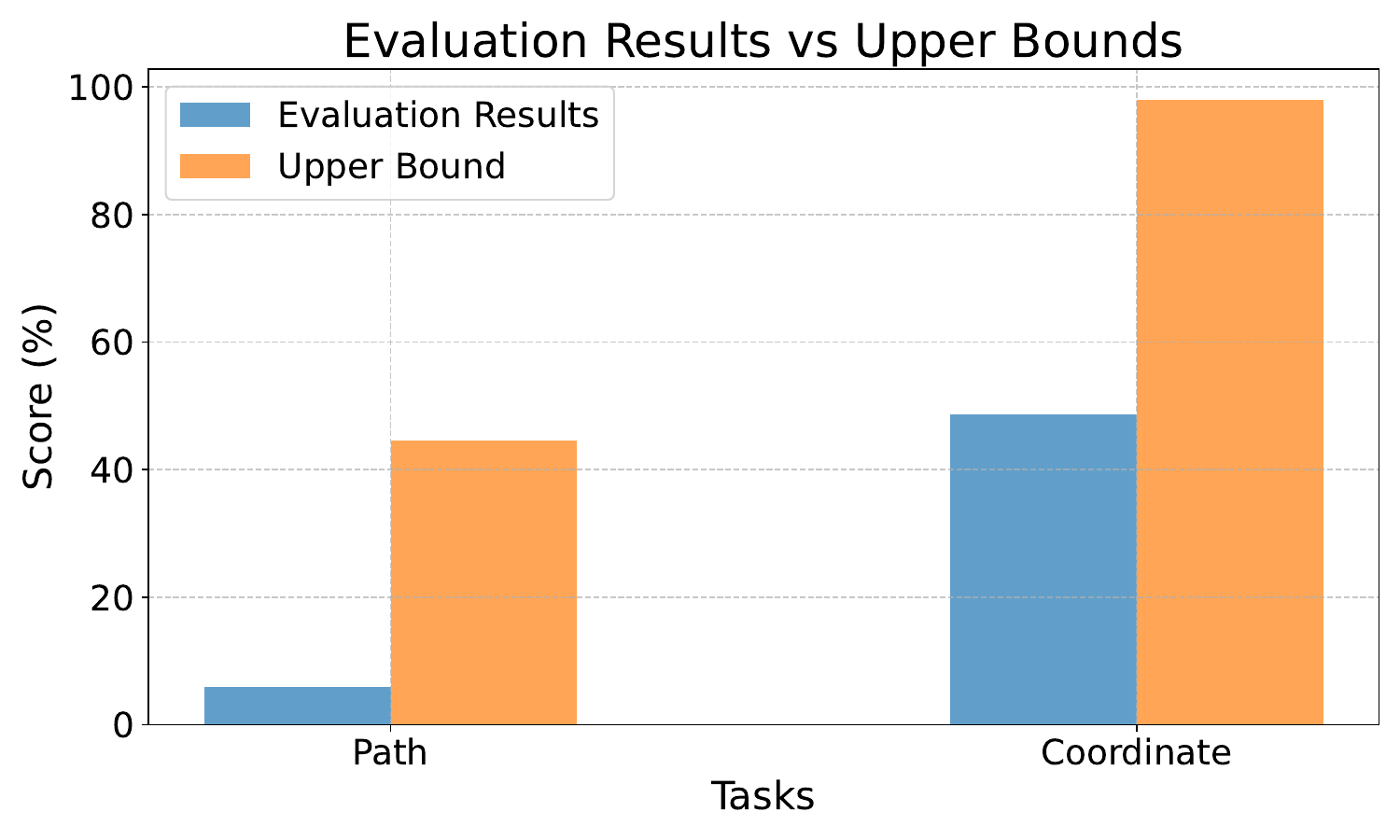}
    \caption{The comparison of benchmark evaluation results with upper bound on coordinate and path tasks, where the upper bound is the testing result of the LVLM fine-tuned on partial benchmark dataset. 
    }
    \label{fig:generative_eval_vs_upper}
     \vspace{-0.8em}
\end{figure}

We find that a more diverse dataset offers limited improvement in model robustness, likely due to insufficient data volume (as exhaustively covering these variations requires many more variants). Moreover, while directly training the model on spatial visual tasks improves its performance on position and direction tasks, it does not enhance robustness against position and orientation variations. As Figure~\ref{fig:generative_eval_vs_upper} suggests, after training, despite improvements in the coordinate task, the path tracing task remains underperforming, due to the lack of sustained visual attention. This reveals that the vulnerability is fundamentally rooted in architectural design choices rather than data limitations. 
\vspace{-1.0em}

The design of the vision encoder utilizes patch-based tokenization~\citep{radford2021learningtransferablevisualmodels} and positional embeddings, which may lead to information fragmentation due to arbitrary patch partitioning and position sensitivity induced by explicit positional encoding. Furthermore, the cascading pipeline architecture amplifies vulnerabilities at each component, underscoring the need for a new architectural approach. Although unified architectures currently underperform on visual understanding tasks due to training stability challenges~\citep{chen2025janusprounifiedmultimodalunderstanding}, this direction remains promising. 

\section{Conclusion}
\vspace{-0.4em}

We present an evaluation framework, V$^2$R-Bench, designed to assess the robustness of LVLMs against visual variations. Our results show significant vulnerabilities in existing LVLMs and identify their origins as twofold: insufficient multimodal alignment and error accumulation inherent in pipeline model architectures. While synthetic data augmentation \cite{qin2025scalinglawssyntheticdata} seems as a mitigation strategy, fundamental advancement in the field requires a shift towards native multimodal architectures rather than the current approach of concatenating separate language and vision modalities. We aim to draw greater attention to the importance of LVLM visual robustness and inspire future research toward more robust architectural paradigms.

\section*{Limitations}
Aligned with the paper track's focus, this work primarily concentrates on identifying a novel problem, establishing corresponding evaluation methods, and providing initial analytical insights into these vulnerabilities. Regarding solutions to the identified vulnerabilities, only two categories of data construction approaches were explored in the automated construction pipeline, while potential improvements through architectural modifications, pre-training strategies, test-time self-correction or self-consistency approaches \cite{he2024selfcorrectionrefinementlearningframework,wang2025calmunleashingcrosslingualselfaligning}, and novel modality alignment methods remain unexplored due to their substantial resource requirements. The detailed analysis presented in this paper aims to provide insights for future research addressing these challenges. Additionally, while this study examines four fundamental types of visual variations, it does not exhaust the infinite possibilities of visual transformations, leaving some long-tail cases unexplored.

\section*{Ethics Statements}
This study reveals a fundamental vulnerability in LVLMs - their lack of robustness to visual variations that naturally occur from camera parameter adjustments and environmental changes. Such robustness deficiency results in significant output inconsistencies across visually similar scenarios, compromising model reliability in real-world deployments.

A critical security implication emerges from these findings: visual variations could serve as a novel attack vector. Through strategic object placement with specific positions, scales, orientations, or contexts, these fundamental visual variations could be exploited to manipulate model behavior. Unlike conventional adversarial attacks requiring sophisticated training procedures, this approach requires no training and generates natural images without artificial artifacts, making such attacks particularly challenging for existing detection mechanisms.

The identified vulnerability underscores the need for increased attention within the research community to these fundamental yet profoundly impactful visual variations, rather than solely pursuing state-of-the-art performance on complex tasks. Enhanced robustness to these variations is crucial for ensuring consistent model performance in natural environments. Moreover, as attacks based on visual variations exploit inherent model vulnerabilities rather than crafted adversarial strategies, addressing this robustness issue becomes essential for improving both the reliability and security of deployed LVLMs.

\bibliography{main}

\begin{thebibliography}{83}
\providecommand{\natexlab}[1]{#1}

\bibitem[{Abdin et~al.(2024)Abdin, Aneja, Awadalla, Awadallah, Awan, Bach, Bahree, Bakhtiari, Bao, Behl, Benhaim, Bilenko, Bjorck, Bubeck, Cai, Cai, Chaudhary, Chen, Chen, Chen, Chen, Chen, Cheng, Chopra, Dai, Dixon, Eldan, Fragoso, Gao, Gao, Gao, Garg, Giorno, Goswami, Gunasekar, Haider, Hao, Hewett, Hu, Huynh, Iter, Jacobs, Javaheripi, Jin, Karampatziakis, Kauffmann, Khademi, Kim, Kim, Kurilenko, Lee, Lee, Li, Li, Liang, Liden, Lin, Lin, Liu, Liu, Liu, Liu, Liu, Luo, Madan, Mahmoudzadeh, Majercak, Mazzola, Mendes, Mitra, Modi, Nguyen, Norick, Patra, Perez-Becker, Portet, Pryzant, Qin, Radmilac, Ren, de~Rosa, Rosset, Roy, Ruwase, Saarikivi, Saied, Salim, Santacroce, Shah, Shang, Sharma, Shen, Shukla, Song, Tanaka, Tupini, Vaddamanu, Wang, Wang, Wang, Wang, Wang, Wang, Ward, Wen, Witte, Wu, Wu, Wyatt, Xiao, Xu, Xu, Xu, Xue, Yadav, Yang, Yang, Yang, Yang, Yu, Yuan, Zhang, Zhang, Zhang, Zhang, Zhang, Zhang, Zhang, and Zhou}]{abdin2024phi3technicalreporthighly}
Marah Abdin, Jyoti Aneja, Hany Awadalla, Ahmed Awadallah, Ammar~Ahmad Awan, Nguyen Bach, Amit Bahree, Arash Bakhtiari, Jianmin Bao, Harkirat Behl, Alon Benhaim, Misha Bilenko, Johan Bjorck, Sébastien Bubeck, Martin Cai, Qin Cai, Vishrav Chaudhary, Dong Chen, Dongdong Chen, Weizhu Chen, Yen-Chun Chen, Yi-Ling Chen, Hao Cheng, Parul Chopra, Xiyang Dai, Matthew Dixon, Ronen Eldan, Victor Fragoso, Jianfeng Gao, Mei Gao, Min Gao, Amit Garg, Allie~Del Giorno, Abhishek Goswami, Suriya Gunasekar, Emman Haider, Junheng Hao, Russell~J. Hewett, Wenxiang Hu, Jamie Huynh, Dan Iter, Sam~Ade Jacobs, Mojan Javaheripi, Xin Jin, Nikos Karampatziakis, Piero Kauffmann, Mahoud Khademi, Dongwoo Kim, Young~Jin Kim, Lev Kurilenko, James~R. Lee, Yin~Tat Lee, Yuanzhi Li, Yunsheng Li, Chen Liang, Lars Liden, Xihui Lin, Zeqi Lin, Ce~Liu, Liyuan Liu, Mengchen Liu, Weishung Liu, Xiaodong Liu, Chong Luo, Piyush Madan, Ali Mahmoudzadeh, David Majercak, Matt Mazzola, Caio César~Teodoro Mendes, Arindam Mitra, Hardik Modi, Anh Nguyen,
  Brandon Norick, Barun Patra, Daniel Perez-Becker, Thomas Portet, Reid Pryzant, Heyang Qin, Marko Radmilac, Liliang Ren, Gustavo de~Rosa, Corby Rosset, Sambudha Roy, Olatunji Ruwase, Olli Saarikivi, Amin Saied, Adil Salim, Michael Santacroce, Shital Shah, Ning Shang, Hiteshi Sharma, Yelong Shen, Swadheen Shukla, Xia Song, Masahiro Tanaka, Andrea Tupini, Praneetha Vaddamanu, Chunyu Wang, Guanhua Wang, Lijuan Wang, Shuohang Wang, Xin Wang, Yu~Wang, Rachel Ward, Wen Wen, Philipp Witte, Haiping Wu, Xiaoxia Wu, Michael Wyatt, Bin Xiao, Can Xu, Jiahang Xu, Weijian Xu, Jilong Xue, Sonali Yadav, Fan Yang, Jianwei Yang, Yifan Yang, Ziyi Yang, Donghan Yu, Lu~Yuan, Chenruidong Zhang, Cyril Zhang, Jianwen Zhang, Li~Lyna Zhang, Yi~Zhang, Yue Zhang, Yunan Zhang, and Xiren Zhou. 2024.
\newblock \href {https://arxiv.org/abs/2404.14219} {Phi-3 technical report: A highly capable language model locally on your phone}.
\newblock \emph{Preprint}, arXiv:2404.14219.

\bibitem[{Alain and Bengio(2018)}]{alain2018understandingintermediatelayersusing}
Guillaume Alain and Yoshua Bengio. 2018.
\newblock \href {https://arxiv.org/abs/1610.01644} {Understanding intermediate layers using linear classifier probes}.
\newblock \emph{Preprint}, arXiv:1610.01644.

\bibitem[{Alayrac et~al.(2022)Alayrac, Donahue, Luc, Miech, Barr, Hasson, Lenc, Mensch, Millican, Reynolds, Ring, Rutherford, Cabi, Han, Gong, Samangooei, Monteiro, Menick, Borgeaud, Brock, Nematzadeh, Sharifzadeh, Binkowski, Barreira, Vinyals, Zisserman, and Simonyan}]{alayrac2022flamingovisuallanguagemodel}
Jean-Baptiste Alayrac, Jeff Donahue, Pauline Luc, Antoine Miech, Iain Barr, Yana Hasson, Karel Lenc, Arthur Mensch, Katie Millican, Malcolm Reynolds, Roman Ring, Eliza Rutherford, Serkan Cabi, Tengda Han, Zhitao Gong, Sina Samangooei, Marianne Monteiro, Jacob Menick, Sebastian Borgeaud, Andrew Brock, Aida Nematzadeh, Sahand Sharifzadeh, Mikolaj Binkowski, Ricardo Barreira, Oriol Vinyals, Andrew Zisserman, and Karen Simonyan. 2022.
\newblock \href {https://arxiv.org/abs/2204.14198} {Flamingo: a visual language model for few-shot learning}.
\newblock \emph{Preprint}, arXiv:2204.14198.

\bibitem[{Ananthram et~al.(2024)Ananthram, Stengel-Eskin, Vondrick, Bansal, and McKeown}]{ananthram2024perspectivediagnosingwesterncultural}
Amith Ananthram, Elias Stengel-Eskin, Carl Vondrick, Mohit Bansal, and Kathleen McKeown. 2024.
\newblock \href {https://arxiv.org/abs/2406.11665} {See it from my perspective: Diagnosing the western cultural bias of large vision-language models in image understanding}.
\newblock \emph{Preprint}, arXiv:2406.11665.

\bibitem[{Arenas(2020)}]{arenas_blender_sdg_2023}
Federico Arenas. 2020.
\newblock \href {https://github.com/federicoarenasl/blender-sdg} {Blender {SDG} {Icons} {Generator}}.
\newblock https://github.com/federicoarenasl/blender-sdg.
\newblock Blender add-on for generating 3D Sustainable Development Goals (SDG) icons.

\bibitem[{Bai et~al.(2023{\natexlab{a}})Bai, Bai, Chu, Cui, Dang, Deng, Fan, Ge, Han, Huang, Hui, Ji, Li, Lin, Lin, Liu, Liu, Lu, Lu, Ma, Men, Ren, Ren, Tan, Tan, Tu, Wang, Wang, Wang, Wu, Xu, Xu, Yang, Yang, Yang, Yang, Yao, Yu, Yuan, Yuan, Zhang, Zhang, Zhang, Zhang, Zhou, Zhou, Zhou, and Zhu}]{qwen}
Jinze Bai, Shuai Bai, Yunfei Chu, Zeyu Cui, Kai Dang, Xiaodong Deng, Yang Fan, Wenbin Ge, Yu~Han, Fei Huang, Binyuan Hui, Luo Ji, Mei Li, Junyang Lin, Runji Lin, Dayiheng Liu, Gao Liu, Chengqiang Lu, Keming Lu, Jianxin Ma, Rui Men, Xingzhang Ren, Xuancheng Ren, Chuanqi Tan, Sinan Tan, Jianhong Tu, Peng Wang, Shijie Wang, Wei Wang, Shengguang Wu, Benfeng Xu, Jin Xu, An~Yang, Hao Yang, Jian Yang, Shusheng Yang, Yang Yao, Bowen Yu, Hongyi Yuan, Zheng Yuan, Jianwei Zhang, Xingxuan Zhang, Yichang Zhang, Zhenru Zhang, Chang Zhou, Jingren Zhou, Xiaohuan Zhou, and Tianhang Zhu. 2023{\natexlab{a}}.
\newblock Qwen technical report.
\newblock \emph{arXiv preprint arXiv:2309.16609}.

\bibitem[{Bai et~al.(2023{\natexlab{b}})Bai, Bai, Yang, Wang, Tan, Wang, Lin, Zhou, and Zhou}]{Qwen-VL}
Jinze Bai, Shuai Bai, Shusheng Yang, Shijie Wang, Sinan Tan, Peng Wang, Junyang Lin, Chang Zhou, and Jingren Zhou. 2023{\natexlab{b}}.
\newblock Qwen-vl: A versatile vision-language model for understanding, localization, text reading, and beyond.
\newblock \emph{arXiv preprint arXiv:2308.12966}.

\bibitem[{Bailey et~al.(2024)Bailey, Ahdritz, Kleiman, Swaroop, Doshi-Velez, and Pan}]{bailey2023soft}
Luke Bailey, Gustaf Ahdritz, Anat Kleiman, Siddharth Swaroop, Finale Doshi-Velez, and Weiwei Pan. 2024.
\newblock Soft prompting might be a bug, not a feature.

\bibitem[{Chen et~al.(2023{\natexlab{a}})Chen, Li, Dong, Zhang, He, Wang, Zhao, and Lin}]{chen2023sharegpt4vimprovinglargemultimodal}
Lin Chen, Jinsong Li, Xiaoyi Dong, Pan Zhang, Conghui He, Jiaqi Wang, Feng Zhao, and Dahua Lin. 2023{\natexlab{a}}.
\newblock \href {https://arxiv.org/abs/2311.12793} {Sharegpt4v: Improving large multi-modal models with better captions}.
\newblock \emph{Preprint}, arXiv:2311.12793.

\bibitem[{Chen et~al.(2024{\natexlab{a}})Chen, Ye, Wang, Li, Deng, Li, Li, Duan, Huang, Su, Wang, Zhang, Fu, Cai, Zhuang, Seibel, He, and Qiao}]{chen2024gmaimmbenchcomprehensivemultimodalevaluation}
Pengcheng Chen, Jin Ye, Guoan Wang, Yanjun Li, Zhongying Deng, Wei Li, Tianbin Li, Haodong Duan, Ziyan Huang, Yanzhou Su, Benyou Wang, Shaoting Zhang, Bin Fu, Jianfei Cai, Bohan Zhuang, Eric~J Seibel, Junjun He, and Yu~Qiao. 2024{\natexlab{a}}.
\newblock \href {https://arxiv.org/abs/2408.03361} {Gmai-mmbench: A comprehensive multimodal evaluation benchmark towards general medical ai}.
\newblock \emph{Preprint}, arXiv:2408.03361.

\bibitem[{Chen et~al.(2023{\natexlab{b}})Chen, Gu, Han, Ma, Torr, and Tresp}]{chen2023benchmarking}
Shuo Chen, Jindong Gu, Zhen Han, Yunpu Ma, Philip Torr, and Volker Tresp. 2023{\natexlab{b}}.
\newblock \href {https://openreview.net/forum?id=4d8dO5sAeM} {Benchmarking robustness of adaptation methods on pre-trained vision-language models}.
\newblock In \emph{Thirty-seventh Conference on Neural Information Processing Systems Datasets and Benchmarks Track}.

\bibitem[{Chen et~al.(2025{\natexlab{a}})Chen, Wu, Liu, Pan, Liu, Xie, Yu, and Ruan}]{chen2025janusprounifiedmultimodalunderstanding}
Xiaokang Chen, Zhiyu Wu, Xingchao Liu, Zizheng Pan, Wen Liu, Zhenda Xie, Xingkai Yu, and Chong Ruan. 2025{\natexlab{a}}.
\newblock \href {https://arxiv.org/abs/2501.17811} {Janus-pro: Unified multimodal understanding and generation with data and model scaling}.
\newblock \emph{Preprint}, arXiv:2501.17811.

\bibitem[{Chen et~al.(2024{\natexlab{b}})Chen, Shi, Lu, He, Qian, Fang, Yin, Ouyang, Shao, Qiao, Lu, and Sheng}]{chen2024rh20tpprimitivelevelroboticdataset}
Zeren Chen, Zhelun Shi, Xiaoya Lu, Lehan He, Sucheng Qian, Hao~Shu Fang, Zhenfei Yin, Wanli Ouyang, Jing Shao, Yu~Qiao, Cewu Lu, and Lu~Sheng. 2024{\natexlab{b}}.
\newblock \href {https://arxiv.org/abs/2403.19622} {Rh20t-p: A primitive-level robotic dataset towards composable generalization agents}.
\newblock \emph{Preprint}, arXiv:2403.19622.

\bibitem[{Chen et~al.(2025{\natexlab{b}})Chen, Wang, Cao, Liu, Gao, Cui, Zhu, Ye, Tian, Liu, Gu, Wang, Li, Ren, Chen, Luo, Wang, Jiang, Wang, He, Shi, Zhang, Lv, Wang, Shao, Chu, Tu, He, Wu, Deng, Ge, Chen, Zhang, Wang, Dou, Lu, Zhu, Lu, Lin, Qiao, Dai, and Wang}]{chen2025expandingperformanceboundariesopensource}
Zhe Chen, Weiyun Wang, Yue Cao, Yangzhou Liu, Zhangwei Gao, Erfei Cui, Jinguo Zhu, Shenglong Ye, Hao Tian, Zhaoyang Liu, Lixin Gu, Xuehui Wang, Qingyun Li, Yimin Ren, Zixuan Chen, Jiapeng Luo, Jiahao Wang, Tan Jiang, Bo~Wang, Conghui He, Botian Shi, Xingcheng Zhang, Han Lv, Yi~Wang, Wenqi Shao, Pei Chu, Zhongying Tu, Tong He, Zhiyong Wu, Huipeng Deng, Jiaye Ge, Kai Chen, Kaipeng Zhang, Limin Wang, Min Dou, Lewei Lu, Xizhou Zhu, Tong Lu, Dahua Lin, Yu~Qiao, Jifeng Dai, and Wenhai Wang. 2025{\natexlab{b}}.
\newblock \href {https://arxiv.org/abs/2412.05271} {Expanding performance boundaries of open-source multimodal models with model, data, and test-time scaling}.
\newblock \emph{Preprint}, arXiv:2412.05271.

\bibitem[{Chen et~al.(2024{\natexlab{c}})Chen, Wu, Wang, Su, Chen, Xing, Zhong, Zhang, Zhu, Lu et~al.}]{chen2024internvl}
Zhe Chen, Jiannan Wu, Wenhai Wang, Weijie Su, Guo Chen, Sen Xing, Muyan Zhong, Qinglong Zhang, Xizhou Zhu, Lewei Lu, et~al. 2024{\natexlab{c}}.
\newblock Internvl: Scaling up vision foundation models and aligning for generic visual-linguistic tasks.
\newblock In \emph{Proceedings of the IEEE/CVF Conference on Computer Vision and Pattern Recognition}, pages 24185--24198.

\bibitem[{Corneanu et~al.(2024)Corneanu, Gadde, and Martinez}]{10483967}
Ciprian Corneanu, Raghudeep Gadde, and Aleix~M Martinez. 2024.
\newblock \href {https://doi.org/10.1109/WACV57701.2024.00428} {Latentpaint: Image inpainting in latent space with diffusion models}.
\newblock In \emph{2024 IEEE/CVF Winter Conference on Applications of Computer Vision (WACV)}, pages 4322--4331.

\bibitem[{Deitke et~al.(2024)Deitke, Clark, Lee, Tripathi, Yang, Park, Salehi, Muennighoff, Lo, Soldaini, Lu, Anderson, Bransom, Ehsani, Ngo, Chen, Patel, Yatskar, Callison-Burch, Head, Hendrix, Bastani, VanderBilt, Lambert, Chou, Chheda, Sparks, Skjonsberg, Schmitz, Sarnat, Bischoff, Walsh, Newell, Wolters, Gupta, Zeng, Borchardt, Groeneveld, Nam, Lebrecht, Wittlif, Schoenick, Michel, Krishna, Weihs, Smith, Hajishirzi, Girshick, Farhadi, and Kembhavi}]{deitke2024molmopixmoopenweights}
Matt Deitke, Christopher Clark, Sangho Lee, Rohun Tripathi, Yue Yang, Jae~Sung Park, Mohammadreza Salehi, Niklas Muennighoff, Kyle Lo, Luca Soldaini, Jiasen Lu, Taira Anderson, Erin Bransom, Kiana Ehsani, Huong Ngo, YenSung Chen, Ajay Patel, Mark Yatskar, Chris Callison-Burch, Andrew Head, Rose Hendrix, Favyen Bastani, Eli VanderBilt, Nathan Lambert, Yvonne Chou, Arnavi Chheda, Jenna Sparks, Sam Skjonsberg, Michael Schmitz, Aaron Sarnat, Byron Bischoff, Pete Walsh, Chris Newell, Piper Wolters, Tanmay Gupta, Kuo-Hao Zeng, Jon Borchardt, Dirk Groeneveld, Crystal Nam, Sophie Lebrecht, Caitlin Wittlif, Carissa Schoenick, Oscar Michel, Ranjay Krishna, Luca Weihs, Noah~A. Smith, Hannaneh Hajishirzi, Ross Girshick, Ali Farhadi, and Aniruddha Kembhavi. 2024.
\newblock \href {https://arxiv.org/abs/2409.17146} {Molmo and pixmo: Open weights and open data for state-of-the-art vision-language models}.
\newblock \emph{Preprint}, arXiv:2409.17146.

\bibitem[{Devlin et~al.(2019)Devlin, Chang, Lee, and Toutanova}]{devlin-etal-2019-bert}
Jacob Devlin, Ming-Wei Chang, Kenton Lee, and Kristina Toutanova. 2019.
\newblock \href {https://doi.org/10.18653/v1/N19-1423} {{BERT}: Pre-training of deep bidirectional transformers for language understanding}.
\newblock In \emph{Proceedings of the 2019 Conference of the North {A}merican Chapter of the Association for Computational Linguistics: Human Language Technologies, Volume 1 (Long and Short Papers)}, pages 4171--4186, Minneapolis, Minnesota. Association for Computational Linguistics.

\bibitem[{Dubey et~al.(2024)Dubey, Jauhri, Pandey, Kadian, Al-Dahle, Letman, Mathur, Schelten, Yang, Fan, Goyal, Hartshorn, Yang, Mitra, Sravankumar, Korenev, Hinsvark, Rao, Zhang, Rodriguez, Gregerson, Spataru, Roziere, Biron, Tang, Chern, Caucheteux, Nayak, Bi, Marra, McConnell, Keller, Touret, Wu, Wong, Ferrer, Nikolaidis, Allonsius, Song, Pintz, Livshits, Esiobu, Choudhary, Mahajan, Garcia-Olano, Perino, Hupkes, Lakomkin, AlBadawy, Lobanova, Dinan, Smith, Radenovic, Zhang, Synnaeve, Lee, Anderson, Nail, Mialon, Pang, Cucurell, Nguyen, Korevaar, Xu, Touvron, Zarov, Ibarra, Kloumann, Misra, Evtimov, Copet, Lee, Geffert, Vranes, Park, Mahadeokar, Shah, van~der Linde, Billock, Hong, Lee, Fu, Chi, Huang, Liu, Wang, Yu, Bitton, Spisak, Park, Rocca, Johnstun, Saxe, Jia, Alwala, Upasani, Plawiak, Li, Heafield, Stone, El-Arini, Iyer, Malik, Chiu, Bhalla, Rantala-Yeary, van~der Maaten, Chen, Tan, Jenkins, Martin, Madaan, Malo, Blecher, Landzaat, de~Oliveira, Muzzi, Pasupuleti, Singh, Paluri, Kardas, Oldham, Rita,
  Pavlova, Kambadur, Lewis, Si, Singh, Hassan, Goyal, Torabi, Bashlykov, Bogoychev, Chatterji, Duchenne, Çelebi, Alrassy, Zhang, Li, Vasic, Weng, Bhargava, Dubal, Krishnan, Koura, Xu, He, Dong, Srinivasan, Ganapathy, Calderer, Cabral, Stojnic, Raileanu, Girdhar, Patel, Sauvestre, Polidoro, Sumbaly, Taylor, Silva, Hou, Wang, Hosseini, Chennabasappa, Singh, Bell, Kim, Edunov, Nie, Narang, Raparthy, Shen, Wan, Bhosale, Zhang, Vandenhende, Batra, Whitman, Sootla, Collot, Gururangan, Borodinsky, Herman, Fowler, Sheasha, Georgiou, Scialom, Speckbacher, Mihaylov, Xiao, Karn, Goswami, Gupta, Ramanathan, Kerkez, Gonguet, Do, Vogeti, Petrovic, Chu, Xiong, Fu, Meers, Martinet, Wang, Tan, Xie, Jia, Wang, Goldschlag, Gaur, Babaei, Wen, Song, Zhang, Li, Mao, Coudert, Yan, Chen, Papakipos, Singh, Grattafiori, Jain, Kelsey, Shajnfeld, Gangidi, Victoria, Goldstand, Menon, Sharma, Boesenberg, Vaughan, Baevski, Feinstein, Kallet, Sangani, Yunus, Lupu, Alvarado, Caples, Gu, Ho, Poulton, Ryan, Ramchandani, Franco, Saraf,
  Chowdhury, Gabriel, Bharambe, Eisenman, Yazdan, James, Maurer, Leonhardi, Huang, Loyd, Paola, Paranjape, Liu, Wu, Ni, Hancock, Wasti, Spence, Stojkovic, Gamido, Montalvo, Parker, Burton, Mejia, Wang, Kim, Zhou, Hu, Chu, Cai, Tindal, Feichtenhofer, Civin, Beaty, Kreymer, Li, Wyatt, Adkins, Xu, Testuggine, David, Parikh, Liskovich, Foss, Wang, Le, Holland, Dowling, Jamil, Montgomery, Presani, Hahn, Wood, Brinkman, Arcaute, Dunbar, Smothers, Sun, Kreuk, Tian, Ozgenel, Caggioni, Guzmán, Kanayet, Seide, Florez, Schwarz, Badeer, Swee, Halpern, Thattai, Herman, Sizov, Guangyi, Zhang, Lakshminarayanan, Shojanazeri, Zou, Wang, Zha, Habeeb, Rudolph, Suk, Aspegren, Goldman, Damlaj, Molybog, Tufanov, Veliche, Gat, Weissman, Geboski, Kohli, Asher, Gaya, Marcus, Tang, Chan, Zhen, Reizenstein, Teboul, Zhong, Jin, Yang, Cummings, Carvill, Shepard, McPhie, Torres, Ginsburg, Wang, Wu, U, Saxena, Prasad, Khandelwal, Zand, Matosich, Veeraraghavan, Michelena, Li, Huang, Chawla, Lakhotia, Huang, Chen, Garg, A, Silva, Bell,
  Zhang, Guo, Yu, Moshkovich, Wehrstedt, Khabsa, Avalani, Bhatt, Tsimpoukelli, Mankus, Hasson, Lennie, Reso, Groshev, Naumov, Lathi, Keneally, Seltzer, Valko, Restrepo, Patel, Vyatskov, Samvelyan, Clark, Macey, Wang, Hermoso, Metanat, Rastegari, Bansal, Santhanam, Parks, White, Bawa, Singhal, Egebo, Usunier, Laptev, Dong, Zhang, Cheng, Chernoguz, Hart, Salpekar, Kalinli, Kent, Parekh, Saab, Balaji, Rittner, Bontrager, Roux, Dollar, Zvyagina, Ratanchandani, Yuvraj, Liang, Alao, Rodriguez, Ayub, Murthy, Nayani, Mitra, Li, Hogan, Battey, Wang, Maheswari, Howes, Rinott, Bondu, Datta, Chugh, Hunt, Dhillon, Sidorov, Pan, Verma, Yamamoto, Ramaswamy, Lindsay, Lindsay, Feng, Lin, Zha, Shankar, Zhang, Zhang, Wang, Agarwal, Sajuyigbe, Chintala, Max, Chen, Kehoe, Satterfield, Govindaprasad, Gupta, Cho, Virk, Subramanian, Choudhury, Goldman, Remez, Glaser, Best, Kohler, Robinson, Li, Zhang, Matthews, Chou, Shaked, Vontimitta, Ajayi, Montanez, Mohan, Kumar, Mangla, Albiero, Ionescu, Poenaru, Mihailescu, Ivanov, Li, Wang,
  Jiang, Bouaziz, Constable, Tang, Wang, Wu, Wang, Xia, Wu, Gao, Chen, Hu, Jia, Qi, Li, Zhang, Zhang, Adi, Nam, Yu, Wang, Hao, Qian, He, Rait, DeVito, Rosnbrick, Wen, Yang, and Zhao}]{dubey2024llama3herdmodels}
Abhimanyu Dubey, Abhinav Jauhri, Abhinav Pandey, Abhishek Kadian, Ahmad Al-Dahle, Aiesha Letman, Akhil Mathur, Alan Schelten, Amy Yang, Angela Fan, Anirudh Goyal, Anthony Hartshorn, Aobo Yang, Archi Mitra, Archie Sravankumar, Artem Korenev, Arthur Hinsvark, Arun Rao, Aston Zhang, Aurelien Rodriguez, Austen Gregerson, Ava Spataru, Baptiste Roziere, Bethany Biron, Binh Tang, Bobbie Chern, Charlotte Caucheteux, Chaya Nayak, Chloe Bi, Chris Marra, Chris McConnell, Christian Keller, Christophe Touret, Chunyang Wu, Corinne Wong, Cristian~Canton Ferrer, Cyrus Nikolaidis, Damien Allonsius, Daniel Song, Danielle Pintz, Danny Livshits, David Esiobu, Dhruv Choudhary, Dhruv Mahajan, Diego Garcia-Olano, Diego Perino, Dieuwke Hupkes, Egor Lakomkin, Ehab AlBadawy, Elina Lobanova, Emily Dinan, Eric~Michael Smith, Filip Radenovic, Frank Zhang, Gabriel Synnaeve, Gabrielle Lee, Georgia~Lewis Anderson, Graeme Nail, Gregoire Mialon, Guan Pang, Guillem Cucurell, Hailey Nguyen, Hannah Korevaar, Hu~Xu, Hugo Touvron, Iliyan Zarov,
  Imanol~Arrieta Ibarra, Isabel Kloumann, Ishan Misra, Ivan Evtimov, Jade Copet, Jaewon Lee, Jan Geffert, Jana Vranes, Jason Park, Jay Mahadeokar, Jeet Shah, Jelmer van~der Linde, Jennifer Billock, Jenny Hong, Jenya Lee, Jeremy Fu, Jianfeng Chi, Jianyu Huang, Jiawen Liu, Jie Wang, Jiecao Yu, Joanna Bitton, Joe Spisak, Jongsoo Park, Joseph Rocca, Joshua Johnstun, Joshua Saxe, Junteng Jia, Kalyan~Vasuden Alwala, Kartikeya Upasani, Kate Plawiak, Ke~Li, Kenneth Heafield, Kevin Stone, Khalid El-Arini, Krithika Iyer, Kshitiz Malik, Kuenley Chiu, Kunal Bhalla, Lauren Rantala-Yeary, Laurens van~der Maaten, Lawrence Chen, Liang Tan, Liz Jenkins, Louis Martin, Lovish Madaan, Lubo Malo, Lukas Blecher, Lukas Landzaat, Luke de~Oliveira, Madeline Muzzi, Mahesh Pasupuleti, Mannat Singh, Manohar Paluri, Marcin Kardas, Mathew Oldham, Mathieu Rita, Maya Pavlova, Melanie Kambadur, Mike Lewis, Min Si, Mitesh~Kumar Singh, Mona Hassan, Naman Goyal, Narjes Torabi, Nikolay Bashlykov, Nikolay Bogoychev, Niladri Chatterji, Olivier
  Duchenne, Onur Çelebi, Patrick Alrassy, Pengchuan Zhang, Pengwei Li, Petar Vasic, Peter Weng, Prajjwal Bhargava, Pratik Dubal, Praveen Krishnan, Punit~Singh Koura, Puxin Xu, Qing He, Qingxiao Dong, Ragavan Srinivasan, Raj Ganapathy, Ramon Calderer, Ricardo~Silveira Cabral, Robert Stojnic, Roberta Raileanu, Rohit Girdhar, Rohit Patel, Romain Sauvestre, Ronnie Polidoro, Roshan Sumbaly, Ross Taylor, Ruan Silva, Rui Hou, Rui Wang, Saghar Hosseini, Sahana Chennabasappa, Sanjay Singh, Sean Bell, Seohyun~Sonia Kim, Sergey Edunov, Shaoliang Nie, Sharan Narang, Sharath Raparthy, Sheng Shen, Shengye Wan, Shruti Bhosale, Shun Zhang, Simon Vandenhende, Soumya Batra, Spencer Whitman, Sten Sootla, Stephane Collot, Suchin Gururangan, Sydney Borodinsky, Tamar Herman, Tara Fowler, Tarek Sheasha, Thomas Georgiou, Thomas Scialom, Tobias Speckbacher, Todor Mihaylov, Tong Xiao, Ujjwal Karn, Vedanuj Goswami, Vibhor Gupta, Vignesh Ramanathan, Viktor Kerkez, Vincent Gonguet, Virginie Do, Vish Vogeti, Vladan Petrovic, Weiwei Chu,
  Wenhan Xiong, Wenyin Fu, Whitney Meers, Xavier Martinet, Xiaodong Wang, Xiaoqing~Ellen Tan, Xinfeng Xie, Xuchao Jia, Xuewei Wang, Yaelle Goldschlag, Yashesh Gaur, Yasmine Babaei, Yi~Wen, Yiwen Song, Yuchen Zhang, Yue Li, Yuning Mao, Zacharie~Delpierre Coudert, Zheng Yan, Zhengxing Chen, Zoe Papakipos, Aaditya Singh, Aaron Grattafiori, Abha Jain, Adam Kelsey, Adam Shajnfeld, Adithya Gangidi, Adolfo Victoria, Ahuva Goldstand, Ajay Menon, Ajay Sharma, Alex Boesenberg, Alex Vaughan, Alexei Baevski, Allie Feinstein, Amanda Kallet, Amit Sangani, Anam Yunus, Andrei Lupu, Andres Alvarado, Andrew Caples, Andrew Gu, Andrew Ho, Andrew Poulton, Andrew Ryan, Ankit Ramchandani, Annie Franco, Aparajita Saraf, Arkabandhu Chowdhury, Ashley Gabriel, Ashwin Bharambe, Assaf Eisenman, Azadeh Yazdan, Beau James, Ben Maurer, Benjamin Leonhardi, Bernie Huang, Beth Loyd, Beto~De Paola, Bhargavi Paranjape, Bing Liu, Bo~Wu, Boyu Ni, Braden Hancock, Bram Wasti, Brandon Spence, Brani Stojkovic, Brian Gamido, Britt Montalvo, Carl
  Parker, Carly Burton, Catalina Mejia, Changhan Wang, Changkyu Kim, Chao Zhou, Chester Hu, Ching-Hsiang Chu, Chris Cai, Chris Tindal, Christoph Feichtenhofer, Damon Civin, Dana Beaty, Daniel Kreymer, Daniel Li, Danny Wyatt, David Adkins, David Xu, Davide Testuggine, Delia David, Devi Parikh, Diana Liskovich, Didem Foss, Dingkang Wang, Duc Le, Dustin Holland, Edward Dowling, Eissa Jamil, Elaine Montgomery, Eleonora Presani, Emily Hahn, Emily Wood, Erik Brinkman, Esteban Arcaute, Evan Dunbar, Evan Smothers, Fei Sun, Felix Kreuk, Feng Tian, Firat Ozgenel, Francesco Caggioni, Francisco Guzmán, Frank Kanayet, Frank Seide, Gabriela~Medina Florez, Gabriella Schwarz, Gada Badeer, Georgia Swee, Gil Halpern, Govind Thattai, Grant Herman, Grigory Sizov, Guangyi, Zhang, Guna Lakshminarayanan, Hamid Shojanazeri, Han Zou, Hannah Wang, Hanwen Zha, Haroun Habeeb, Harrison Rudolph, Helen Suk, Henry Aspegren, Hunter Goldman, Ibrahim Damlaj, Igor Molybog, Igor Tufanov, Irina-Elena Veliche, Itai Gat, Jake Weissman, James
  Geboski, James Kohli, Japhet Asher, Jean-Baptiste Gaya, Jeff Marcus, Jeff Tang, Jennifer Chan, Jenny Zhen, Jeremy Reizenstein, Jeremy Teboul, Jessica Zhong, Jian Jin, Jingyi Yang, Joe Cummings, Jon Carvill, Jon Shepard, Jonathan McPhie, Jonathan Torres, Josh Ginsburg, Junjie Wang, Kai Wu, Kam~Hou U, Karan Saxena, Karthik Prasad, Kartikay Khandelwal, Katayoun Zand, Kathy Matosich, Kaushik Veeraraghavan, Kelly Michelena, Keqian Li, Kun Huang, Kunal Chawla, Kushal Lakhotia, Kyle Huang, Lailin Chen, Lakshya Garg, Lavender A, Leandro Silva, Lee Bell, Lei Zhang, Liangpeng Guo, Licheng Yu, Liron Moshkovich, Luca Wehrstedt, Madian Khabsa, Manav Avalani, Manish Bhatt, Maria Tsimpoukelli, Martynas Mankus, Matan Hasson, Matthew Lennie, Matthias Reso, Maxim Groshev, Maxim Naumov, Maya Lathi, Meghan Keneally, Michael~L. Seltzer, Michal Valko, Michelle Restrepo, Mihir Patel, Mik Vyatskov, Mikayel Samvelyan, Mike Clark, Mike Macey, Mike Wang, Miquel~Jubert Hermoso, Mo~Metanat, Mohammad Rastegari, Munish Bansal, Nandhini
  Santhanam, Natascha Parks, Natasha White, Navyata Bawa, Nayan Singhal, Nick Egebo, Nicolas Usunier, Nikolay~Pavlovich Laptev, Ning Dong, Ning Zhang, Norman Cheng, Oleg Chernoguz, Olivia Hart, Omkar Salpekar, Ozlem Kalinli, Parkin Kent, Parth Parekh, Paul Saab, Pavan Balaji, Pedro Rittner, Philip Bontrager, Pierre Roux, Piotr Dollar, Polina Zvyagina, Prashant Ratanchandani, Pritish Yuvraj, Qian Liang, Rachad Alao, Rachel Rodriguez, Rafi Ayub, Raghotham Murthy, Raghu Nayani, Rahul Mitra, Raymond Li, Rebekkah Hogan, Robin Battey, Rocky Wang, Rohan Maheswari, Russ Howes, Ruty Rinott, Sai~Jayesh Bondu, Samyak Datta, Sara Chugh, Sara Hunt, Sargun Dhillon, Sasha Sidorov, Satadru Pan, Saurabh Verma, Seiji Yamamoto, Sharadh Ramaswamy, Shaun Lindsay, Shaun Lindsay, Sheng Feng, Shenghao Lin, Shengxin~Cindy Zha, Shiva Shankar, Shuqiang Zhang, Shuqiang Zhang, Sinong Wang, Sneha Agarwal, Soji Sajuyigbe, Soumith Chintala, Stephanie Max, Stephen Chen, Steve Kehoe, Steve Satterfield, Sudarshan Govindaprasad, Sumit Gupta,
  Sungmin Cho, Sunny Virk, Suraj Subramanian, Sy~Choudhury, Sydney Goldman, Tal Remez, Tamar Glaser, Tamara Best, Thilo Kohler, Thomas Robinson, Tianhe Li, Tianjun Zhang, Tim Matthews, Timothy Chou, Tzook Shaked, Varun Vontimitta, Victoria Ajayi, Victoria Montanez, Vijai Mohan, Vinay~Satish Kumar, Vishal Mangla, Vítor Albiero, Vlad Ionescu, Vlad Poenaru, Vlad~Tiberiu Mihailescu, Vladimir Ivanov, Wei Li, Wenchen Wang, Wenwen Jiang, Wes Bouaziz, Will Constable, Xiaocheng Tang, Xiaofang Wang, Xiaojian Wu, Xiaolan Wang, Xide Xia, Xilun Wu, Xinbo Gao, Yanjun Chen, Ye~Hu, Ye~Jia, Ye~Qi, Yenda Li, Yilin Zhang, Ying Zhang, Yossi Adi, Youngjin Nam, Yu, Wang, Yuchen Hao, Yundi Qian, Yuzi He, Zach Rait, Zachary DeVito, Zef Rosnbrick, Zhaoduo Wen, Zhenyu Yang, and Zhiwei Zhao. 2024.
\newblock \href {https://arxiv.org/abs/2407.21783} {The llama 3 herd of models}.
\newblock \emph{Preprint}, arXiv:2407.21783.

\bibitem[{Fei et~al.(2024)Fei, Yao, Zhang, Liu, Zhang, and Chua}]{fei-etal-2024-multimodal}
Hao Fei, Yuan Yao, Zhuosheng Zhang, Fuxiao Liu, Ao~Zhang, and Tat-Seng Chua. 2024.
\newblock \href {https://aclanthology.org/2024.lrec-tutorials.1/} {From multimodal {LLM} to human-level {AI}: Modality, instruction, reasoning, efficiency and beyond}.
\newblock In \emph{Proceedings of the 2024 Joint International Conference on Computational Linguistics, Language Resources and Evaluation (LREC-COLING 2024): Tutorial Summaries}, pages 1--8, Torino, Italia. ELRA and ICCL.

\bibitem[{Fu et~al.(2024{\natexlab{a}})Fu, Chen, Shen, Qin, Zhang, Lin, Yang, Zheng, Li, Sun, Wu, and Ji}]{fu2024mmecomprehensiveevaluationbenchmark}
Chaoyou Fu, Peixian Chen, Yunhang Shen, Yulei Qin, Mengdan Zhang, Xu~Lin, Jinrui Yang, Xiawu Zheng, Ke~Li, Xing Sun, Yunsheng Wu, and Rongrong Ji. 2024{\natexlab{a}}.
\newblock \href {https://arxiv.org/abs/2306.13394} {Mme: A comprehensive evaluation benchmark for multimodal large language models}.
\newblock \emph{Preprint}, arXiv:2306.13394.

\bibitem[{Fu et~al.(2024{\natexlab{b}})Fu, Hu, Li, Feng, Wang, Lin, Roth, Smith, Ma, and Krishna}]{fu2024blinkmultimodallargelanguage}
Xingyu Fu, Yushi Hu, Bangzheng Li, Yu~Feng, Haoyu Wang, Xudong Lin, Dan Roth, Noah~A. Smith, Wei-Chiu Ma, and Ranjay Krishna. 2024{\natexlab{b}}.
\newblock \href {https://arxiv.org/abs/2404.12390} {Blink: Multimodal large language models can see but not perceive}.
\newblock \emph{Preprint}, arXiv:2404.12390.

\bibitem[{Galib et~al.(2024)Galib, Wang, Xu, Pfeiffer, Chesler, Landry, and Ambati}]{galib2024h2ovlmississippivisionlanguagemodels}
Shaikat Galib, Shanshan Wang, Guanshuo Xu, Pascal Pfeiffer, Ryan Chesler, Mark Landry, and Sri~Satish Ambati. 2024.
\newblock \href {https://arxiv.org/abs/2410.13611} {H2ovl-mississippi vision language models technical report}.
\newblock \emph{Preprint}, arXiv:2410.13611.

\bibitem[{GLM et~al.(2024)GLM, Zeng, Xu, Wang, Zhang, Yin, Rojas, Feng, Zhao, Lai, Yu, Wang, Sun, Zhang, Cheng, Gui, Tang, Zhang, Li, Zhao, Wu, Zhong, Liu, Huang, Zhang, Zheng, Lu, Duan, Zhang, Cao, Yang, Tam, Zhao, Liu, Xia, Zhang, Gu, Lv, Liu, Liu, Yang, Song, Zhang, An, Xu, Niu, Yang, Li, Bai, Dong, Qi, Wang, Yang, Du, Hou, and Wang}]{glm2024chatglm}
Team GLM, Aohan Zeng, Bin Xu, Bowen Wang, Chenhui Zhang, Da~Yin, Diego Rojas, Guanyu Feng, Hanlin Zhao, Hanyu Lai, Hao Yu, Hongning Wang, Jiadai Sun, Jiajie Zhang, Jiale Cheng, Jiayi Gui, Jie Tang, Jing Zhang, Juanzi Li, Lei Zhao, Lindong Wu, Lucen Zhong, Mingdao Liu, Minlie Huang, Peng Zhang, Qinkai Zheng, Rui Lu, Shuaiqi Duan, Shudan Zhang, Shulin Cao, Shuxun Yang, Weng~Lam Tam, Wenyi Zhao, Xiao Liu, Xiao Xia, Xiaohan Zhang, Xiaotao Gu, Xin Lv, Xinghan Liu, Xinyi Liu, Xinyue Yang, Xixuan Song, Xunkai Zhang, Yifan An, Yifan Xu, Yilin Niu, Yuantao Yang, Yueyan Li, Yushi Bai, Yuxiao Dong, Zehan Qi, Zhaoyu Wang, Zhen Yang, Zhengxiao Du, Zhenyu Hou, and Zihan Wang. 2024.
\newblock \href {https://arxiv.org/abs/2406.12793} {Chatglm: A family of large language models from glm-130b to glm-4 all tools}.
\newblock \emph{Preprint}, arXiv:2406.12793.

\bibitem[{Gu et~al.(2022)Gu, Han, Liu, and Huang}]{gu-etal-2022-ppt}
Yuxian Gu, Xu~Han, Zhiyuan Liu, and Minlie Huang. 2022.
\newblock \href {https://doi.org/10.18653/v1/2022.acl-long.576} {{PPT}: Pre-trained prompt tuning for few-shot learning}.
\newblock In \emph{Proceedings of the 60th Annual Meeting of the Association for Computational Linguistics (Volume 1: Long Papers)}, pages 8410--8423, Dublin, Ireland. Association for Computational Linguistics.

\bibitem[{He et~al.(2024)He, Lin, Wang, Fung, and Ji}]{he2024selfcorrectionrefinementlearningframework}
Jiayi He, Hehai Lin, Qingyun Wang, Yi~Fung, and Heng Ji. 2024.
\newblock \href {https://arxiv.org/abs/2410.04055} {Self-correction is more than refinement: A learning framework for visual and language reasoning tasks}.
\newblock \emph{Preprint}, arXiv:2410.04055.

\bibitem[{He et~al.(2025)He, Polisetty, Fan, Huang, Wu, and Fung}]{he2025mmboundaryadvancingmllmknowledge}
Zhitao He, Sandeep Polisetty, Zhiyuan Fan, Yuchen Huang, Shujin Wu, and Yi~R. Fung. 2025.
\newblock \href {https://arxiv.org/abs/2505.23224} {Mmboundary: Advancing mllm knowledge boundary awareness through reasoning step confidence calibration}.
\newblock \emph{Preprint}, arXiv:2505.23224.

\bibitem[{Howard et~al.(2024)Howard, Bhiwandiwalla, Fraser, and Kiritchenko}]{howard2024uncoveringbiaslargevisionlanguage}
Phillip Howard, Anahita Bhiwandiwalla, Kathleen~C. Fraser, and Svetlana Kiritchenko. 2024.
\newblock \href {https://arxiv.org/abs/2404.00166} {Uncovering bias in large vision-language models with counterfactuals}.
\newblock \emph{Preprint}, arXiv:2404.00166.

\bibitem[{Hu et~al.(2024)Hu, Li, Lu, Shao, He, Qiao, and Luo}]{hu2024omnimedvqanewlargescalecomprehensive}
Yutao Hu, Tianbin Li, Quanfeng Lu, Wenqi Shao, Junjun He, Yu~Qiao, and Ping Luo. 2024.
\newblock \href {https://arxiv.org/abs/2402.09181} {Omnimedvqa: A new large-scale comprehensive evaluation benchmark for medical lvlm}.
\newblock \emph{Preprint}, arXiv:2402.09181.

\bibitem[{Huang et~al.(2025)Huang, Chan, Fung, Qiu, Zhou, Joty, Chang, and Ji}]{10787102}
Kung-Hsiang Huang, Hou~Pong Chan, May Fung, Haoyi Qiu, Mingyang Zhou, Shafiq Joty, Shih-Fu Chang, and Heng Ji. 2025.
\newblock \href {https://doi.org/10.1109/TKDE.2024.3513320} {From pixels to insights: A survey on automatic chart understanding in the era of large foundation models}.
\newblock \emph{IEEE Transactions on Knowledge and Data Engineering}, 37(5):2550--2568.

\bibitem[{Huang et~al.(2024)Huang, Zhou, Chan, Fung, Wang, Zhang, Chang, and Ji}]{huang-etal-2024-lvlms}
Kung-Hsiang Huang, Mingyang Zhou, Hou~Pong Chan, Yi~Fung, Zhenhailong Wang, Lingyu Zhang, Shih-Fu Chang, and Heng Ji. 2024.
\newblock \href {https://doi.org/10.18653/v1/2024.findings-acl.41} {Do {LVLM}s understand charts? analyzing and correcting factual errors in chart captioning}.
\newblock In \emph{Findings of the Association for Computational Linguistics: ACL 2024}, pages 730--749, Bangkok, Thailand. Association for Computational Linguistics.

\bibitem[{Jiang et~al.(2023)Jiang, Sablayrolles, Mensch, Bamford, Chaplot, de~las Casas, Bressand, Lengyel, Lample, Saulnier, Lavaud, Lachaux, Stock, Scao, Lavril, Wang, Lacroix, and Sayed}]{jiang2023mistral7b}
Albert~Q. Jiang, Alexandre Sablayrolles, Arthur Mensch, Chris Bamford, Devendra~Singh Chaplot, Diego de~las Casas, Florian Bressand, Gianna Lengyel, Guillaume Lample, Lucile Saulnier, Lélio~Renard Lavaud, Marie-Anne Lachaux, Pierre Stock, Teven~Le Scao, Thibaut Lavril, Thomas Wang, Timothée Lacroix, and William~El Sayed. 2023.
\newblock \href {https://arxiv.org/abs/2310.06825} {Mistral 7b}.
\newblock \emph{Preprint}, arXiv:2310.06825.

\bibitem[{Jung et~al.(2019)Jung, Kang, Mentch, and Hovy}]{jung-etal-2019-earlier}
Taehee Jung, Dongyeop Kang, Lucas Mentch, and Eduard Hovy. 2019.
\newblock \href {https://doi.org/10.18653/v1/D19-1327} {Earlier isn`t always better: Sub-aspect analysis on corpus and system biases in summarization}.
\newblock In \emph{Proceedings of the 2019 Conference on Empirical Methods in Natural Language Processing and the 9th International Joint Conference on Natural Language Processing (EMNLP-IJCNLP)}, pages 3324--3335, Hong Kong, China. Association for Computational Linguistics.

\bibitem[{Koo et~al.(2024)Koo, Lee, Raheja, Park, Kim, and Kang}]{koo2024benchmarkingcognitivebiaseslarge}
Ryan Koo, Minhwa Lee, Vipul Raheja, Jong~Inn Park, Zae~Myung Kim, and Dongyeop Kang. 2024.
\newblock \href {https://arxiv.org/abs/2309.17012} {Benchmarking cognitive biases in large language models as evaluators}.
\newblock \emph{Preprint}, arXiv:2309.17012.

\bibitem[{Lee et~al.(2024)Lee, Tu, Wong, Zheng, Zhou, Mai, Roberts, Yasunaga, Yao, Xie, and Liang}]{lee2024vhelmholisticevaluationvision}
Tony Lee, Haoqin Tu, Chi~Heem Wong, Wenhao Zheng, Yiyang Zhou, Yifan Mai, Josselin~Somerville Roberts, Michihiro Yasunaga, Huaxiu Yao, Cihang Xie, and Percy Liang. 2024.
\newblock \href {https://arxiv.org/abs/2410.07112} {Vhelm: A holistic evaluation of vision language models}.
\newblock \emph{Preprint}, arXiv:2410.07112.

\bibitem[{Lester et~al.(2021)Lester, Al-Rfou, and Constant}]{lester2021powerscaleparameterefficientprompt}
Brian Lester, Rami Al-Rfou, and Noah Constant. 2021.
\newblock \href {https://arxiv.org/abs/2104.08691} {The power of scale for parameter-efficient prompt tuning}.
\newblock \emph{Preprint}, arXiv:2104.08691.

\bibitem[{Li et~al.(2024)Li, Zhang, Guo, Zhang, Li, Zhang, Zhang, Zhang, Li, Liu, and Li}]{li2024llavaonevisioneasyvisualtask}
Bo~Li, Yuanhan Zhang, Dong Guo, Renrui Zhang, Feng Li, Hao Zhang, Kaichen Zhang, Peiyuan Zhang, Yanwei Li, Ziwei Liu, and Chunyuan Li. 2024.
\newblock \href {https://arxiv.org/abs/2408.03326} {Llava-onevision: Easy visual task transfer}.
\newblock \emph{Preprint}, arXiv:2408.03326.

\bibitem[{Li et~al.(2023{\natexlab{a}})Li, Ge, Ge, Wang, Wang, Zhang, and Shan}]{li2023seedbench2benchmarkingmultimodallarge}
Bohao Li, Yuying Ge, Yixiao Ge, Guangzhi Wang, Rui Wang, Ruimao Zhang, and Ying Shan. 2023{\natexlab{a}}.
\newblock \href {https://arxiv.org/abs/2311.17092} {Seed-bench-2: Benchmarking multimodal large language models}.
\newblock \emph{Preprint}, arXiv:2311.17092.

\bibitem[{Li et~al.(2023{\natexlab{b}})Li, Li, Savarese, and Hoi}]{li2023blip2bootstrappinglanguageimagepretraining}
Junnan Li, Dongxu Li, Silvio Savarese, and Steven Hoi. 2023{\natexlab{b}}.
\newblock \href {https://arxiv.org/abs/2301.12597} {Blip-2: Bootstrapping language-image pre-training with frozen image encoders and large language models}.
\newblock \emph{Preprint}, arXiv:2301.12597.

\bibitem[{Li et~al.(2023{\natexlab{c}})Li, Yin, Li, Chen, Wang, Ren, Li, Yang, Xu, Sun, Kong, and Liu}]{li2023m3itlargescaledatasetmultimodal}
Lei Li, Yuwei Yin, Shicheng Li, Liang Chen, Peiyi Wang, Shuhuai Ren, Mukai Li, Yazheng Yang, Jingjing Xu, Xu~Sun, Lingpeng Kong, and Qi~Liu. 2023{\natexlab{c}}.
\newblock \href {https://arxiv.org/abs/2306.04387} {M$^3$it: A large-scale dataset towards multi-modal multilingual instruction tuning}.
\newblock \emph{Preprint}, arXiv:2306.04387.

\bibitem[{Liang et~al.(2024)Liang, Goindani, Chafekar, Mathur, Yu, Salakhutdinov, and Morency}]{liang2024hemmholisticevaluationmultimodal}
Paul~Pu Liang, Akshay Goindani, Talha Chafekar, Leena Mathur, Haofei Yu, Ruslan Salakhutdinov, and Louis-Philippe Morency. 2024.
\newblock \href {https://arxiv.org/abs/2407.03418} {Hemm: Holistic evaluation of multimodal foundation models}.
\newblock \emph{Preprint}, arXiv:2407.03418.

\bibitem[{Liu et~al.(2024{\natexlab{a}})Liu, Cai, Zhou, Qu, Fang, Sun, and Hu}]{liu2024are}
Daizong Liu, Xiaowen Cai, Pan Zhou, Xiaoye Qu, Xiang Fang, Lichao Sun, and Wei Hu. 2024{\natexlab{a}}.
\newblock \href {https://openreview.net/forum?id=q8XGHj7yrC} {Are large vision-language models robust to adversarial visual transformations?}

\bibitem[{Liu et~al.(2024{\natexlab{b}})Liu, Yang, Qu, Zhou, Fang, Tang, Wan, and Sun}]{liu2024pandoras}
Daizong Liu, Mingyu Yang, Xiaoye Qu, Pan Zhou, Xiang Fang, Keke Tang, Yao Wan, and Lichao Sun. 2024{\natexlab{b}}.
\newblock \href {https://openreview.net/forum?id=gDpWYpocE1} {Pandora's box: Towards building universal attackers against real-world large vision-language models}.
\newblock In \emph{The Thirty-eighth Annual Conference on Neural Information Processing Systems}.

\bibitem[{Liu et~al.(2024{\natexlab{c}})Liu, Li, Li, Li, Zhang, Shen, and Lee}]{liu2024llavanext}
Haotian Liu, Chunyuan Li, Yuheng Li, Bo~Li, Yuanhan Zhang, Sheng Shen, and Yong~Jae Lee. 2024{\natexlab{c}}.
\newblock \href {https://llava-vl.github.io/blog/2024-01-30-llava-next/} {Llava-next: Improved reasoning, ocr, and world knowledge}.

\bibitem[{Liu et~al.(2023)Liu, Li, Wu, and Lee}]{liu2023visualinstructiontuning}
Haotian Liu, Chunyuan Li, Qingyang Wu, and Yong~Jae Lee. 2023.
\newblock \href {https://arxiv.org/abs/2304.08485} {Visual instruction tuning}.
\newblock \emph{Preprint}, arXiv:2304.08485.

\bibitem[{Liu et~al.(2022)Liu, Ji, Fu, Tam, Du, Yang, and Tang}]{liu-etal-2022-p}
Xiao Liu, Kaixuan Ji, Yicheng Fu, Weng Tam, Zhengxiao Du, Zhilin Yang, and Jie Tang. 2022.
\newblock \href {https://doi.org/10.18653/v1/2022.acl-short.8} {{P}-tuning: Prompt tuning can be comparable to fine-tuning across scales and tasks}.
\newblock In \emph{Proceedings of the 60th Annual Meeting of the Association for Computational Linguistics (Volume 2: Short Papers)}, pages 61--68, Dublin, Ireland. Association for Computational Linguistics.

\bibitem[{Liu et~al.(2024{\natexlab{d}})Liu, Duan, Zhang, Li, Zhang, Zhao, Yuan, Wang, He, Liu, Chen, and Lin}]{liu2024mmbenchmultimodalmodelallaround}
Yuan Liu, Haodong Duan, Yuanhan Zhang, Bo~Li, Songyang Zhang, Wangbo Zhao, Yike Yuan, Jiaqi Wang, Conghui He, Ziwei Liu, Kai Chen, and Dahua Lin. 2024{\natexlab{d}}.
\newblock \href {https://arxiv.org/abs/2307.06281} {Mmbench: Is your multi-modal model an all-around player?}
\newblock \emph{Preprint}, arXiv:2307.06281.

\bibitem[{Lu et~al.(2024{\natexlab{a}})Lu, Liu, Zhang, Wang, Dong, Liu, Sun, Ren, Li, Yang, Sun, Deng, Xu, Xie, and Ruan}]{lu2024deepseekvlrealworldvisionlanguageunderstanding}
Haoyu Lu, Wen Liu, Bo~Zhang, Bingxuan Wang, Kai Dong, Bo~Liu, Jingxiang Sun, Tongzheng Ren, Zhuoshu Li, Hao Yang, Yaofeng Sun, Chengqi Deng, Hanwei Xu, Zhenda Xie, and Chong Ruan. 2024{\natexlab{a}}.
\newblock \href {https://arxiv.org/abs/2403.05525} {Deepseek-vl: Towards real-world vision-language understanding}.
\newblock \emph{Preprint}, arXiv:2403.05525.

\bibitem[{Lu et~al.(2024{\natexlab{b}})Lu, Bansal, Xia, Liu, Li, Hajishirzi, Cheng, Chang, Galley, and Gao}]{lu2024mathvistaevaluatingmathematicalreasoning}
Pan Lu, Hritik Bansal, Tony Xia, Jiacheng Liu, Chunyuan Li, Hannaneh Hajishirzi, Hao Cheng, Kai-Wei Chang, Michel Galley, and Jianfeng Gao. 2024{\natexlab{b}}.
\newblock \href {https://arxiv.org/abs/2310.02255} {Mathvista: Evaluating mathematical reasoning of foundation models in visual contexts}.
\newblock \emph{Preprint}, arXiv:2310.02255.

\bibitem[{Lugmayr et~al.(2022)Lugmayr, Danelljan, Romero, Yu, Timofte, and Gool}]{lugmayr2022repaintinpaintingusingdenoising}
Andreas Lugmayr, Martin Danelljan, Andres Romero, Fisher Yu, Radu Timofte, and Luc~Van Gool. 2022.
\newblock \href {https://arxiv.org/abs/2201.09865} {Repaint: Inpainting using denoising diffusion probabilistic models}.
\newblock \emph{Preprint}, arXiv:2201.09865.

\bibitem[{Luo et~al.(2024{\natexlab{a}})Luo, Yang, Dou, Wang, Liu, Dai, Qiao, and Zhu}]{luo2024monointernvlpushingboundariesmonolithic}
Gen Luo, Xue Yang, Wenhan Dou, Zhaokai Wang, Jiawen Liu, Jifeng Dai, Yu~Qiao, and Xizhou Zhu. 2024{\natexlab{a}}.
\newblock \href {https://arxiv.org/abs/2410.08202} {Mono-internvl: Pushing the boundaries of monolithic multimodal large language models with endogenous visual pre-training}.
\newblock \emph{Preprint}, arXiv:2410.08202.

\bibitem[{Luo et~al.(2024{\natexlab{b}})Luo, Ma, Liu, Guo, and Xiao}]{luo2024jailbreakvbenchmarkassessingrobustness}
Weidi Luo, Siyuan Ma, Xiaogeng Liu, Xiaoyu Guo, and Chaowei Xiao. 2024{\natexlab{b}}.
\newblock \href {https://arxiv.org/abs/2404.03027} {Jailbreakv: A benchmark for assessing the robustness of multimodal large language models against jailbreak attacks}.
\newblock \emph{Preprint}, arXiv:2404.03027.

\bibitem[{Luo et~al.(2017)Luo, Li, Urtasun, and Zemel}]{luo2017understandingeffectivereceptivefield}
Wenjie Luo, Yujia Li, Raquel Urtasun, and Richard Zemel. 2017.
\newblock \href {https://arxiv.org/abs/1701.04128} {Understanding the effective receptive field in deep convolutional neural networks}.
\newblock \emph{Preprint}, arXiv:1701.04128.

\bibitem[{OpenAI et~al.(2024)OpenAI, Achiam, Adler, Agarwal, Ahmad, Akkaya, Aleman, Almeida, Altenschmidt, Altman, Anadkat, Avila, Babuschkin, Balaji, Balcom, Baltescu, Bao, Bavarian, Belgum, Bello, Berdine, Bernadett-Shapiro, Berner, Bogdonoff, Boiko, Boyd, Brakman, Brockman, Brooks, Brundage, Button, Cai, Campbell, Cann, Carey, Carlson, Carmichael, Chan, Chang, Chantzis, Chen, Chen, Chen, Chen, Chen, Chess, Cho, Chu, Chung, Cummings, Currier, Dai, Decareaux, Degry, Deutsch, Deville, Dhar, Dohan, Dowling, Dunning, Ecoffet, Eleti, Eloundou, Farhi, Fedus, Felix, Fishman, Forte, Fulford, Gao, Georges, Gibson, Goel, Gogineni, Goh, Gontijo-Lopes, Gordon, Grafstein, Gray, Greene, Gross, Gu, Guo, Hallacy, Han, Harris, He, Heaton, Heidecke, Hesse, Hickey, Hickey, Hoeschele, Houghton, Hsu, Hu, Hu, Huizinga, Jain, Jain, Jang, Jiang, Jiang, Jin, Jin, Jomoto, Jonn, Jun, Kaftan, Łukasz Kaiser, Kamali, Kanitscheider, Keskar, Khan, Kilpatrick, Kim, Kim, Kim, Kirchner, Kiros, Knight, Kokotajlo, Łukasz Kondraciuk,
  Kondrich, Konstantinidis, Kosic, Krueger, Kuo, Lampe, Lan, Lee, Leike, Leung, Levy, Li, Lim, Lin, Lin, Litwin, Lopez, Lowe, Lue, Makanju, Malfacini, Manning, Markov, Markovski, Martin, Mayer, Mayne, McGrew, McKinney, McLeavey, McMillan, McNeil, Medina, Mehta, Menick, Metz, Mishchenko, Mishkin, Monaco, Morikawa, Mossing, Mu, Murati, Murk, Mély, Nair, Nakano, Nayak, Neelakantan, Ngo, Noh, Ouyang, O'Keefe, Pachocki, Paino, Palermo, Pantuliano, Parascandolo, Parish, Parparita, Passos, Pavlov, Peng, Perelman, de~Avila Belbute~Peres, Petrov, de~Oliveira~Pinto, Michael, Pokorny, Pokrass, Pong, Powell, Power, Power, Proehl, Puri, Radford, Rae, Ramesh, Raymond, Real, Rimbach, Ross, Rotsted, Roussez, Ryder, Saltarelli, Sanders, Santurkar, Sastry, Schmidt, Schnurr, Schulman, Selsam, Sheppard, Sherbakov, Shieh, Shoker, Shyam, Sidor, Sigler, Simens, Sitkin, Slama, Sohl, Sokolowsky, Song, Staudacher, Such, Summers, Sutskever, Tang, Tezak, Thompson, Tillet, Tootoonchian, Tseng, Tuggle, Turley, Tworek, Uribe, Vallone,
  Vijayvergiya, Voss, Wainwright, Wang, Wang, Wang, Ward, Wei, Weinmann, Welihinda, Welinder, Weng, Weng, Wiethoff, Willner, Winter, Wolrich, Wong, Workman, Wu, Wu, Wu, Xiao, Xu, Yoo, Yu, Yuan, Zaremba, Zellers, Zhang, Zhang, Zhao, Zheng, Zhuang, Zhuk, and Zoph}]{openai2024gpt4technicalreport}
OpenAI, Josh Achiam, Steven Adler, Sandhini Agarwal, Lama Ahmad, Ilge Akkaya, Florencia~Leoni Aleman, Diogo Almeida, Janko Altenschmidt, Sam Altman, Shyamal Anadkat, Red Avila, Igor Babuschkin, Suchir Balaji, Valerie Balcom, Paul Baltescu, Haiming Bao, Mohammad Bavarian, Jeff Belgum, Irwan Bello, Jake Berdine, Gabriel Bernadett-Shapiro, Christopher Berner, Lenny Bogdonoff, Oleg Boiko, Madelaine Boyd, Anna-Luisa Brakman, Greg Brockman, Tim Brooks, Miles Brundage, Kevin Button, Trevor Cai, Rosie Campbell, Andrew Cann, Brittany Carey, Chelsea Carlson, Rory Carmichael, Brooke Chan, Che Chang, Fotis Chantzis, Derek Chen, Sully Chen, Ruby Chen, Jason Chen, Mark Chen, Ben Chess, Chester Cho, Casey Chu, Hyung~Won Chung, Dave Cummings, Jeremiah Currier, Yunxing Dai, Cory Decareaux, Thomas Degry, Noah Deutsch, Damien Deville, Arka Dhar, David Dohan, Steve Dowling, Sheila Dunning, Adrien Ecoffet, Atty Eleti, Tyna Eloundou, David Farhi, Liam Fedus, Niko Felix, Simón~Posada Fishman, Juston Forte, Isabella Fulford, Leo
  Gao, Elie Georges, Christian Gibson, Vik Goel, Tarun Gogineni, Gabriel Goh, Rapha Gontijo-Lopes, Jonathan Gordon, Morgan Grafstein, Scott Gray, Ryan Greene, Joshua Gross, Shixiang~Shane Gu, Yufei Guo, Chris Hallacy, Jesse Han, Jeff Harris, Yuchen He, Mike Heaton, Johannes Heidecke, Chris Hesse, Alan Hickey, Wade Hickey, Peter Hoeschele, Brandon Houghton, Kenny Hsu, Shengli Hu, Xin Hu, Joost Huizinga, Shantanu Jain, Shawn Jain, Joanne Jang, Angela Jiang, Roger Jiang, Haozhun Jin, Denny Jin, Shino Jomoto, Billie Jonn, Heewoo Jun, Tomer Kaftan, Łukasz Kaiser, Ali Kamali, Ingmar Kanitscheider, Nitish~Shirish Keskar, Tabarak Khan, Logan Kilpatrick, Jong~Wook Kim, Christina Kim, Yongjik Kim, Jan~Hendrik Kirchner, Jamie Kiros, Matt Knight, Daniel Kokotajlo, Łukasz Kondraciuk, Andrew Kondrich, Aris Konstantinidis, Kyle Kosic, Gretchen Krueger, Vishal Kuo, Michael Lampe, Ikai Lan, Teddy Lee, Jan Leike, Jade Leung, Daniel Levy, Chak~Ming Li, Rachel Lim, Molly Lin, Stephanie Lin, Mateusz Litwin, Theresa Lopez, Ryan
  Lowe, Patricia Lue, Anna Makanju, Kim Malfacini, Sam Manning, Todor Markov, Yaniv Markovski, Bianca Martin, Katie Mayer, Andrew Mayne, Bob McGrew, Scott~Mayer McKinney, Christine McLeavey, Paul McMillan, Jake McNeil, David Medina, Aalok Mehta, Jacob Menick, Luke Metz, Andrey Mishchenko, Pamela Mishkin, Vinnie Monaco, Evan Morikawa, Daniel Mossing, Tong Mu, Mira Murati, Oleg Murk, David Mély, Ashvin Nair, Reiichiro Nakano, Rajeev Nayak, Arvind Neelakantan, Richard Ngo, Hyeonwoo Noh, Long Ouyang, Cullen O'Keefe, Jakub Pachocki, Alex Paino, Joe Palermo, Ashley Pantuliano, Giambattista Parascandolo, Joel Parish, Emy Parparita, Alex Passos, Mikhail Pavlov, Andrew Peng, Adam Perelman, Filipe de~Avila Belbute~Peres, Michael Petrov, Henrique~Ponde de~Oliveira~Pinto, Michael, Pokorny, Michelle Pokrass, Vitchyr~H. Pong, Tolly Powell, Alethea Power, Boris Power, Elizabeth Proehl, Raul Puri, Alec Radford, Jack Rae, Aditya Ramesh, Cameron Raymond, Francis Real, Kendra Rimbach, Carl Ross, Bob Rotsted, Henri Roussez,
  Nick Ryder, Mario Saltarelli, Ted Sanders, Shibani Santurkar, Girish Sastry, Heather Schmidt, David Schnurr, John Schulman, Daniel Selsam, Kyla Sheppard, Toki Sherbakov, Jessica Shieh, Sarah Shoker, Pranav Shyam, Szymon Sidor, Eric Sigler, Maddie Simens, Jordan Sitkin, Katarina Slama, Ian Sohl, Benjamin Sokolowsky, Yang Song, Natalie Staudacher, Felipe~Petroski Such, Natalie Summers, Ilya Sutskever, Jie Tang, Nikolas Tezak, Madeleine~B. Thompson, Phil Tillet, Amin Tootoonchian, Elizabeth Tseng, Preston Tuggle, Nick Turley, Jerry Tworek, Juan Felipe~Cerón Uribe, Andrea Vallone, Arun Vijayvergiya, Chelsea Voss, Carroll Wainwright, Justin~Jay Wang, Alvin Wang, Ben Wang, Jonathan Ward, Jason Wei, CJ~Weinmann, Akila Welihinda, Peter Welinder, Jiayi Weng, Lilian Weng, Matt Wiethoff, Dave Willner, Clemens Winter, Samuel Wolrich, Hannah Wong, Lauren Workman, Sherwin Wu, Jeff Wu, Michael Wu, Kai Xiao, Tao Xu, Sarah Yoo, Kevin Yu, Qiming Yuan, Wojciech Zaremba, Rowan Zellers, Chong Zhang, Marvin Zhang, Shengjia
  Zhao, Tianhao Zheng, Juntang Zhuang, William Zhuk, and Barret Zoph. 2024.
\newblock \href {https://arxiv.org/abs/2303.08774} {Gpt-4 technical report}.
\newblock \emph{Preprint}, arXiv:2303.08774.

\bibitem[{Oquab et~al.(2024)Oquab, Darcet, Moutakanni, Vo, Szafraniec, Khalidov, Fernandez, Haziza, Massa, El-Nouby, Assran, Ballas, Galuba, Howes, Huang, Li, Misra, Rabbat, Sharma, Synnaeve, Xu, Jegou, Mairal, Labatut, Joulin, and Bojanowski}]{oquab2024dinov2learningrobustvisual}
Maxime Oquab, Timothée Darcet, Théo Moutakanni, Huy Vo, Marc Szafraniec, Vasil Khalidov, Pierre Fernandez, Daniel Haziza, Francisco Massa, Alaaeldin El-Nouby, Mahmoud Assran, Nicolas Ballas, Wojciech Galuba, Russell Howes, Po-Yao Huang, Shang-Wen Li, Ishan Misra, Michael Rabbat, Vasu Sharma, Gabriel Synnaeve, Hu~Xu, Hervé Jegou, Julien Mairal, Patrick Labatut, Armand Joulin, and Piotr Bojanowski. 2024.
\newblock \href {https://arxiv.org/abs/2304.07193} {Dinov2: Learning robust visual features without supervision}.
\newblock \emph{Preprint}, arXiv:2304.07193.

\bibitem[{Qin et~al.(2025)Qin, Dong, Zhang, Dong, Huang, Yang, Khademi, Zhang, Awadalla, Fung, Chen, Cheng, and Wei}]{qin2025scalinglawssyntheticdata}
Zeyu Qin, Qingxiu Dong, Xingxing Zhang, Li~Dong, Xiaolong Huang, Ziyi Yang, Mahmoud Khademi, Dongdong Zhang, Hany~Hassan Awadalla, Yi~R. Fung, Weizhu Chen, Minhao Cheng, and Furu Wei. 2025.
\newblock \href {https://arxiv.org/abs/2503.19551} {Scaling laws of synthetic data for language models}.
\newblock \emph{Preprint}, arXiv:2503.19551.

\bibitem[{Radford et~al.(2021)Radford, Kim, Hallacy, Ramesh, Goh, Agarwal, Sastry, Askell, Mishkin, Clark, Krueger, and Sutskever}]{radford2021learningtransferablevisualmodels}
Alec Radford, Jong~Wook Kim, Chris Hallacy, Aditya Ramesh, Gabriel Goh, Sandhini Agarwal, Girish Sastry, Amanda Askell, Pamela Mishkin, Jack Clark, Gretchen Krueger, and Ilya Sutskever. 2021.
\newblock \href {https://arxiv.org/abs/2103.00020} {Learning transferable visual models from natural language supervision}.
\newblock \emph{Preprint}, arXiv:2103.00020.

\bibitem[{Raghu et~al.(2022)Raghu, Unterthiner, Kornblith, Zhang, and Dosovitskiy}]{raghu2022visiontransformerslikeconvolutional}
Maithra Raghu, Thomas Unterthiner, Simon Kornblith, Chiyuan Zhang, and Alexey Dosovitskiy. 2022.
\newblock \href {https://arxiv.org/abs/2108.08810} {Do vision transformers see like convolutional neural networks?}
\newblock \emph{Preprint}, arXiv:2108.08810.

\bibitem[{Steed and Caliskan(2021)}]{10.1145/3442188.3445932}
Ryan Steed and Aylin Caliskan. 2021.
\newblock \href {https://doi.org/10.1145/3442188.3445932} {Image representations learned with unsupervised pre-training contain human-like biases}.
\newblock In \emph{Proceedings of the 2021 ACM Conference on Fairness, Accountability, and Transparency}, FAccT '21, page 701–713, New York, NY, USA. Association for Computing Machinery.

\bibitem[{Team(2024)}]{qvq-72b-preview}
Qwen Team. 2024.
\newblock \href {https://qwenlm.github.io/blog/qvq-72b-preview/} {Qvq: To see the world with wisdom}.

\bibitem[{Touvron et~al.(2023)Touvron, Lavril, Izacard, Martinet, Lachaux, Lacroix, Rozière, Goyal, Hambro, Azhar, Rodriguez, Joulin, Grave, and Lample}]{touvron2023llamaopenefficientfoundation}
Hugo Touvron, Thibaut Lavril, Gautier Izacard, Xavier Martinet, Marie-Anne Lachaux, Timothée Lacroix, Baptiste Rozière, Naman Goyal, Eric Hambro, Faisal Azhar, Aurelien Rodriguez, Armand Joulin, Edouard Grave, and Guillaume Lample. 2023.
\newblock \href {https://arxiv.org/abs/2302.13971} {Llama: Open and efficient foundation language models}.
\newblock \emph{Preprint}, arXiv:2302.13971.

\bibitem[{Wang et~al.(2023{\natexlab{a}})Wang, Liang, Meng, Sun, Shi, Li, Xu, Qu, and Zhou}]{wang2023chatgptgoodnlgevaluator}
Jiaan Wang, Yunlong Liang, Fandong Meng, Zengkui Sun, Haoxiang Shi, Zhixu Li, Jinan Xu, Jianfeng Qu, and Jie Zhou. 2023{\natexlab{a}}.
\newblock \href {https://arxiv.org/abs/2303.04048} {Is chatgpt a good nlg evaluator? a preliminary study}.
\newblock \emph{Preprint}, arXiv:2303.04048.

\bibitem[{Wang et~al.(2023{\natexlab{b}})Wang, Meng, Weng, He, Wu, and Jiang}]{wang2023believepromptinggpt4vbetter}
Junke Wang, Lingchen Meng, Zejia Weng, Bo~He, Zuxuan Wu, and Yu-Gang Jiang. 2023{\natexlab{b}}.
\newblock \href {https://arxiv.org/abs/2311.07574} {To see is to believe: Prompting gpt-4v for better visual instruction tuning}.
\newblock \emph{Preprint}, arXiv:2311.07574.

\bibitem[{Wang et~al.(2024{\natexlab{a}})Wang, Xu, Ye, Yan, Shen, Zhang, Huang, and Sang}]{wang2024mobileagentautonomousmultimodalmobile}
Junyang Wang, Haiyang Xu, Jiabo Ye, Ming Yan, Weizhou Shen, Ji~Zhang, Fei Huang, and Jitao Sang. 2024{\natexlab{a}}.
\newblock \href {https://arxiv.org/abs/2401.16158} {Mobile-agent: Autonomous multi-modal mobile device agent with visual perception}.
\newblock \emph{Preprint}, arXiv:2401.16158.

\bibitem[{Wang et~al.(2024{\natexlab{b}})Wang, Bai, Tan, Wang, Fan, Bai, Chen, Liu, Wang, Ge, Fan, Dang, Du, Ren, Men, Liu, Zhou, Zhou, and Lin}]{wang2024qwen2vlenhancingvisionlanguagemodels}
Peng Wang, Shuai Bai, Sinan Tan, Shijie Wang, Zhihao Fan, Jinze Bai, Keqin Chen, Xuejing Liu, Jialin Wang, Wenbin Ge, Yang Fan, Kai Dang, Mengfei Du, Xuancheng Ren, Rui Men, Dayiheng Liu, Chang Zhou, Jingren Zhou, and Junyang Lin. 2024{\natexlab{b}}.
\newblock \href {https://arxiv.org/abs/2409.12191} {Qwen2-vl: Enhancing vision-language model's perception of the world at any resolution}.
\newblock \emph{Preprint}, arXiv:2409.12191.

\bibitem[{Wang et~al.(2024{\natexlab{c}})Wang, Cao, Zhang, Yuan, Shan, Chen, and Gao}]{wang2024vlbiasbenchcomprehensivebenchmarkevaluating}
Sibo Wang, Xiangkui Cao, Jie Zhang, Zheng Yuan, Shiguang Shan, Xilin Chen, and Wen Gao. 2024{\natexlab{c}}.
\newblock \href {https://arxiv.org/abs/2406.14194} {Vlbiasbench: A comprehensive benchmark for evaluating bias in large vision-language model}.
\newblock \emph{Preprint}, arXiv:2406.14194.

\bibitem[{Wang et~al.(2023{\natexlab{c}})Wang, Mao, Zhu, Xu, Lyu, Li, Chen, Zhang, Chen, Xue, Liu, Lu, Lin, and Pang}]{wang2023embodiedscanholisticmultimodal3d}
Tai Wang, Xiaohan Mao, Chenming Zhu, Runsen Xu, Ruiyuan Lyu, Peisen Li, Xiao Chen, Wenwei Zhang, Kai Chen, Tianfan Xue, Xihui Liu, Cewu Lu, Dahua Lin, and Jiangmiao Pang. 2023{\natexlab{c}}.
\newblock \href {https://arxiv.org/abs/2312.16170} {Embodiedscan: A holistic multi-modal 3d perception suite towards embodied ai}.
\newblock \emph{Preprint}, arXiv:2312.16170.

\bibitem[{Wang et~al.(2025)Wang, Fan, Wang, Fung, and Ji}]{wang2025calmunleashingcrosslingualselfaligning}
Yumeng Wang, Zhiyuan Fan, Qingyun Wang, May Fung, and Heng Ji. 2025.
\newblock \href {https://arxiv.org/abs/2501.18457} {Calm: Unleashing the cross-lingual self-aligning ability of language model question answering}.
\newblock \emph{Preprint}, arXiv:2501.18457.

\bibitem[{Wu and Xie(2023)}]{wu2023vguidedvisualsearch}
Penghao Wu and Saining Xie. 2023.
\newblock \href {https://arxiv.org/abs/2312.14135} {V*: Guided visual search as a core mechanism in multimodal llms}.
\newblock \emph{Preprint}, arXiv:2312.14135.

\bibitem[{Wu et~al.(2024{\natexlab{a}})Wu, Fung, Li, Wan, Chang, and Ji}]{wu2024macaroon}
Shujin Wu, Yi~Fung, Sha Li, Yixin Wan, Kai-Wei Chang, and Heng Ji. 2024{\natexlab{a}}.
\newblock Macaroon: Training vision-language models to be your engaged partners.
\newblock In \emph{Findings of the Association for Computational Linguistics: EMNLP 2024}, pages 7715--7731.

\bibitem[{Wu et~al.(2024{\natexlab{b}})Wu, Xian, Guan, Liang, Chakraborty, Liu, Sadler, Manocha, and Bedi}]{wu2024on}
Xiyang Wu, Ruiqi Xian, Tianrui Guan, Jing Liang, Souradip Chakraborty, Fuxiao Liu, Brian~M. Sadler, Dinesh Manocha, and Amrit Bedi. 2024{\natexlab{b}}.
\newblock \href {https://openreview.net/forum?id=4FpuOMoxsX} {On the safety concerns of deploying {LLM}s/{VLM}s in robotics: Highlighting the risks and vulnerabilities}.
\newblock In \emph{First Vision and Language for Autonomous Driving and Robotics Workshop}.

\bibitem[{Xia et~al.(2024)Xia, Chen, Tian, Gong, Hou, Xu, Wu, Fan, Zhou, Zhu, Zheng, Wang, Wang, Zhang, Bansal, Niethammer, Huang, Zhu, Li, Sun, Ge, Li, Zou, and Yao}]{xia2024carescomprehensivebenchmarktrustworthiness}
Peng Xia, Ze~Chen, Juanxi Tian, Yangrui Gong, Ruibo Hou, Yue Xu, Zhenbang Wu, Zhiyuan Fan, Yiyang Zhou, Kangyu Zhu, Wenhao Zheng, Zhaoyang Wang, Xiao Wang, Xuchao Zhang, Chetan Bansal, Marc Niethammer, Junzhou Huang, Hongtu Zhu, Yun Li, Jimeng Sun, Zongyuan Ge, Gang Li, James Zou, and Huaxiu Yao. 2024.
\newblock \href {https://arxiv.org/abs/2406.06007} {Cares: A comprehensive benchmark of trustworthiness in medical vision language models}.
\newblock \emph{Preprint}, arXiv:2406.06007.

\bibitem[{Yang et~al.(2023)Yang, Song, Li, Zhao, Ge, Li, and Shan}]{yang2023gpt4toolsteachinglargelanguage}
Rui Yang, Lin Song, Yanwei Li, Sijie Zhao, Yixiao Ge, Xiu Li, and Ying Shan. 2023.
\newblock \href {https://arxiv.org/abs/2305.18752} {Gpt4tools: Teaching large language model to use tools via self-instruction}.
\newblock \emph{Preprint}, arXiv:2305.18752.

\bibitem[{Yao et~al.(2024)Yao, Yu, Zhang, Wang, Cui, Zhu, Cai, Li, Zhao, He, Chen, Zhou, Zou, Zhang, Hu, Zheng, Zhou, Cai, Han, Zeng, Li, Liu, and Sun}]{yao2024minicpmvgpt4vlevelmllm}
Yuan Yao, Tianyu Yu, Ao~Zhang, Chongyi Wang, Junbo Cui, Hongji Zhu, Tianchi Cai, Haoyu Li, Weilin Zhao, Zhihui He, Qianyu Chen, Huarong Zhou, Zhensheng Zou, Haoye Zhang, Shengding Hu, Zhi Zheng, Jie Zhou, Jie Cai, Xu~Han, Guoyang Zeng, Dahai Li, Zhiyuan Liu, and Maosong Sun. 2024.
\newblock \href {https://arxiv.org/abs/2408.01800} {Minicpm-v: A gpt-4v level mllm on your phone}.
\newblock \emph{Preprint}, arXiv:2408.01800.

\bibitem[{Yu et~al.(2024)Yu, Yang, Li, Wang, Lin, Liu, Wang, and Wang}]{yu2024mmvetevaluatinglargemultimodal}
Weihao Yu, Zhengyuan Yang, Linjie Li, Jianfeng Wang, Kevin Lin, Zicheng Liu, Xinchao Wang, and Lijuan Wang. 2024.
\newblock \href {https://arxiv.org/abs/2308.02490} {Mm-vet: Evaluating large multimodal models for integrated capabilities}.
\newblock \emph{Preprint}, arXiv:2308.02490.

\bibitem[{Zhang et~al.(2023)Zhang, Yang, Liu, Han, Chen, Huang, Fu, and Yu}]{zhang2023appagentmultimodalagentssmartphone}
Chi Zhang, Zhao Yang, Jiaxuan Liu, Yucheng Han, Xin Chen, Zebiao Huang, Bin Fu, and Gang Yu. 2023.
\newblock \href {https://arxiv.org/abs/2312.13771} {Appagent: Multimodal agents as smartphone users}.
\newblock \emph{Preprint}, arXiv:2312.13771.

\bibitem[{Zhang et~al.(2024{\natexlab{a}})Zhang, Shao, Liu, Ma, Luo, Qiao, Zheng, and Zhang}]{zhang2024bavibenchevaluatingrobustnesslarge}
Hao Zhang, Wenqi Shao, Hong Liu, Yongqiang Ma, Ping Luo, Yu~Qiao, Nanning Zheng, and Kaipeng Zhang. 2024{\natexlab{a}}.
\newblock \href {https://arxiv.org/abs/2403.09346} {B-avibench: Towards evaluating the robustness of large vision-language model on black-box adversarial visual-instructions}.
\newblock \emph{Preprint}, arXiv:2403.09346.

\bibitem[{Zhang et~al.(2025)Zhang, Yao, Pi, Liang, and Fung}]{zhang2025vlm2benchcloserlookvlms}
Jianshu Zhang, Dongyu Yao, Renjie Pi, Paul~Pu Liang, and Yi~R. Fung. 2025.
\newblock \href {https://arxiv.org/abs/2502.12084} {Vlm2-bench: A closer look at how well vlms implicitly link explicit matching visual cues}.
\newblock \emph{Preprint}, arXiv:2502.12084.

\bibitem[{Zhang et~al.(2024{\natexlab{b}})Zhang, Zhang, Tian, Fu, Zhang, Wu, Li, Wang, Wen, Zhang, Wang, Jin, and Tan}]{zhang2024mmerealworldmultimodalllmchallenge}
Yi-Fan Zhang, Huanyu Zhang, Haochen Tian, Chaoyou Fu, Shuangqing Zhang, Junfei Wu, Feng Li, Kun Wang, Qingsong Wen, Zhang Zhang, Liang Wang, Rong Jin, and Tieniu Tan. 2024{\natexlab{b}}.
\newblock \href {https://arxiv.org/abs/2408.13257} {Mme-realworld: Could your multimodal llm challenge high-resolution real-world scenarios that are difficult for humans?}
\newblock \emph{Preprint}, arXiv:2408.13257.

\bibitem[{Zhao et~al.(2023)Zhao, Wu, He, and Huang}]{zhao2023svitscalingvisualinstruction}
Bo~Zhao, Boya Wu, Muyang He, and Tiejun Huang. 2023.
\newblock \href {https://arxiv.org/abs/2307.04087} {Svit: Scaling up visual instruction tuning}.
\newblock \emph{Preprint}, arXiv:2307.04087.

\bibitem[{Zheng et~al.(2023)Zheng, Chiang, Sheng, Zhuang, Wu, Zhuang, Lin, Li, Li, Xing, Zhang, Gonzalez, and Stoica}]{zheng2023judgingllmasajudgemtbenchchatbot}
Lianmin Zheng, Wei-Lin Chiang, Ying Sheng, Siyuan Zhuang, Zhanghao Wu, Yonghao Zhuang, Zi~Lin, Zhuohan Li, Dacheng Li, Eric~P. Xing, Hao Zhang, Joseph~E. Gonzalez, and Ion Stoica. 2023.
\newblock \href {https://arxiv.org/abs/2306.05685} {Judging llm-as-a-judge with mt-bench and chatbot arena}.
\newblock \emph{Preprint}, arXiv:2306.05685.

\bibitem[{Zhou et~al.(2024)Zhou, Fan, Cheng, Yang, Chen, Cui, Wang, Li, Zhang, and Yao}]{zhou2024calibratedselfrewardingvisionlanguage}
Yiyang Zhou, Zhiyuan Fan, Dongjie Cheng, Sihan Yang, Zhaorun Chen, Chenhang Cui, Xiyao Wang, Yun Li, Linjun Zhang, and Huaxiu Yao. 2024.
\newblock \href {https://arxiv.org/abs/2405.14622} {Calibrated self-rewarding vision language models}.
\newblock \emph{Preprint}, arXiv:2405.14622.

\bibitem[{Zhu et~al.(2023)Zhu, Chen, Shen, Li, and Elhoseiny}]{zhu2023minigpt4enhancingvisionlanguageunderstanding}
Deyao Zhu, Jun Chen, Xiaoqian Shen, Xiang Li, and Mohamed Elhoseiny. 2023.
\newblock \href {https://arxiv.org/abs/2304.10592} {Minigpt-4: Enhancing vision-language understanding with advanced large language models}.
\newblock \emph{Preprint}, arXiv:2304.10592.

\end{thebibliography}

\appendix


\section{Discussion}
\label{app:discussion}

\begin{figure*}[t]
    \centering
    \includegraphics[trim={0cm, 2.4cm 0cm 1.18cm},width=1\linewidth]{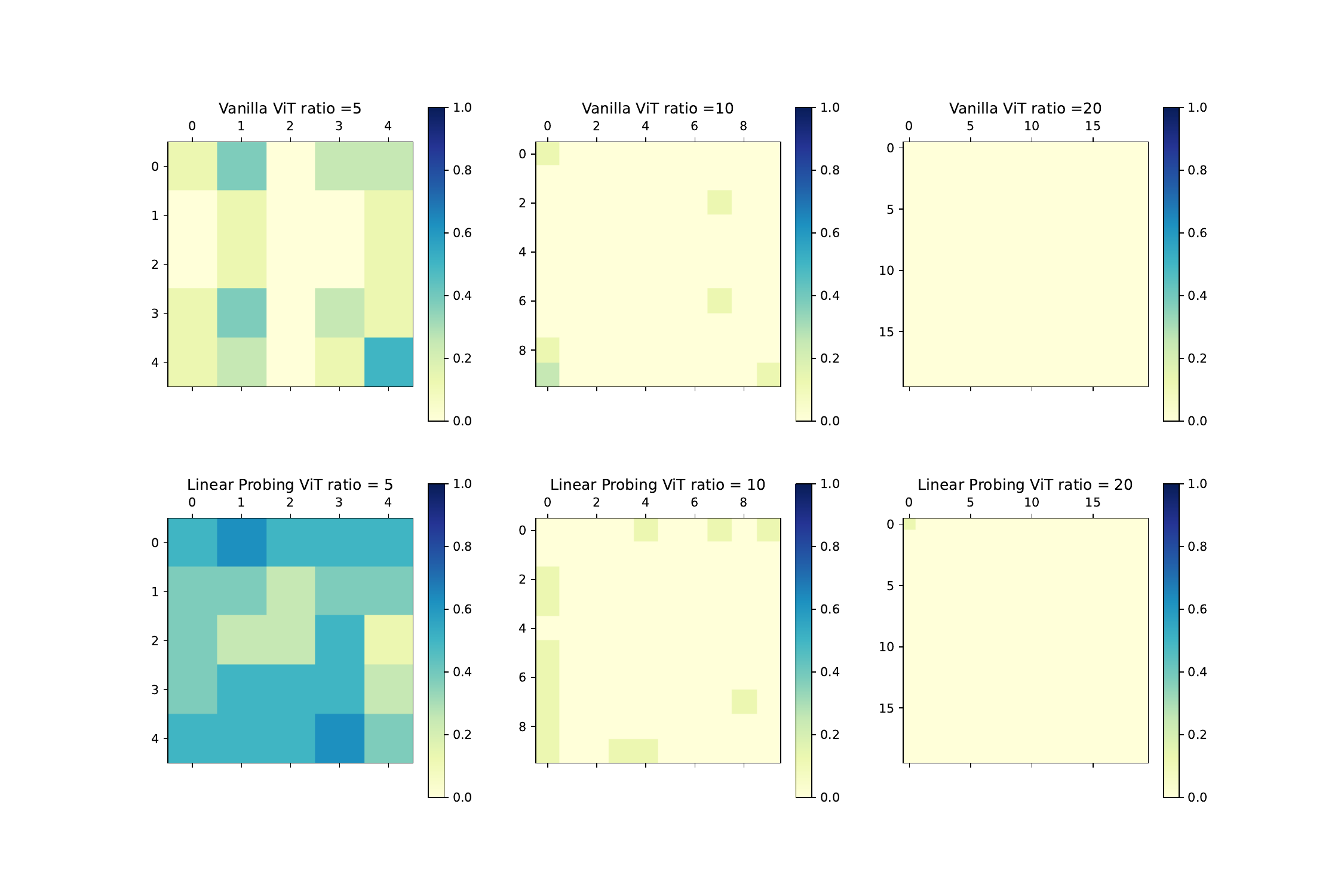}
    \caption{Linear probing results for Vision Encoder show position bias patterns similar to those observed in LVLMs.}
    \label{fig:direction-heat}
\end{figure*}

\paragraph{Summary of Empirical Findings} 
Through the evaluation of V$^2$R-Bench, we discover that introducing visual variation causes inconsistent and unstable output. Such vulnerability is due to the error accumulation in the data generation pipeline. Figure~\ref{fig:direction-heat} reveals that ViT has the tendency to depend on the context during inference rather than truly recognize a target object, which is similar to LVLM. Through component analysis, we claim that multimodal projector is the main cause of the vulnerability to visual variations. The inadequate multimodal alignment causes visual information loss and the decoding result of aligned feature does not form coherent natural language.


\paragraph{The Use of Synthetic Data}
While our framework enables the incorporation of real-world benchmarks, we acknowledge that generated variations may not fully capture the complexity and diversity of real-world visual inputs. To address this, future work can be done to further integrate our framework with diffusion-based contextual blending or 3D synthetic data generation technique such as Blender-SDG~\cite{arenas_blender_sdg_2023}.

\paragraph{Future Directions} V$^2$R-Bench fills a crucial gap in existing evaluations by systematically testing robustness to fundamental visual variations (position, scale, orientation, context)—ubiquitous in real-world scenarios but overlooked by current benchmarks. For instance, autonomous vehicles require consistent object recognition regardless of camera angles, and medical imaging tools must identify anomalies across scales. By integrating existing benchmarks and evaluating with our synthetic data, our framework directly addresses these needs by enabling models to be stress-tested under realistic conditions. Researchers can also easily expand tests to new variations (e.g., lighting changes) or domains (e.g., CT image).

The benchmark’s component-level analysis offers fine-grained diagnostic information, guiding researchers toward targeted improvements. A key future direction lies in enhancing the multimodal alignment mechanisms. Additionally, the identified vulnerabilities are not merely the result of limited data or training strategies but also stem from fundamental architectural constraints requiring further investigation. Moreover, in the longer horizon, we believe that new techniques for enhancing LVLM visual variation robustness can help address the broader objectives of multimodal knowledge boundary awareness and hallucination mitigation as well \cite{he2025mmboundaryadvancingmllmknowledge}.

\section{Prompt Template}

\begin{tcolorbox}[
    colback=blue!3!white,
    colframe=blue!30!white, 
    title=Prompt For Evaluation,
    fonttitle=\bfseries,
    boxrule=0.5pt,
    arc=4pt,
    boxsep=5pt,
    left=6pt,
    right=6pt,
    top=6pt,
    bottom=6pt,
    coltitle=blue!50!black
]
\begin{itemize}
    \item \textbf{Object:} Identify the object in the image.
    \item \textbf{Direction:} List the direction the arrow is pointing in the image using one of the following: up, down, left, right, top-left, bottom-left, top-right, or bottom-right.
    \item \textbf{Coordinate:} This is a coordinate plot with a single point. Provide the coordinate in the format $(x,)$ for 1D, $(x, y)$ for 2D, or $(x, y, z)$ for 3D.
    \item \textbf{Path:} Describe the coordinates of each point along the line from the start to the end in the format $[(x_1, y_1), (x_2, y_2), \dots, (x_n, y_n)]$.
\end{itemize}

\end{tcolorbox}

\begin{table*}[htbp]
\centering
\scalebox{0.9}{
\begin{tabular}{lcccccccccc}
\toprule
\multirow{2}{*}{\textbf{Model}} & \multicolumn{5}{c}{\textbf{Visual Dependency}} & \multicolumn{5}{c}{\textbf{Knowledge Dependency}} \\
\cmidrule(lr){2-6} \cmidrule(lr){7-11}
& Text & B0 & B1 & B2 & B3 & Text & B0 & B1 & B2 & B3 \\
\midrule
GPT-4o & 0.2109 & 0.0475 & 0.0586 & 0.2916 & 0.4724 & 0.0821 & 0.2044 & 0.2030 & 0.2004 & 0.2232 \\
Qwen2-VL & 0.2531 & 0.1169 & 0.2417 & 0.3710 & 0.5344 & 0.2674 & 0.2344 & 0.2284 & 0.3154 & 0.4945 \\
LLaVA-1.6 & 0.2129 & 0.5732 & 1.8404 & 1.8026 & 1.4573 & 0.3743 & 0.5676 & 1.8013 & 1.7919 & 1.4485 \\
\bottomrule
\end{tabular}
}
\caption{Performance Comparison of Models under Different Blur Conditions: LVLMs favor contextual inference over input-output consistency, while LLMs maintain better alignment with inputs. As blur levels increase, LVLMs confidently continue contextual reasoning, implying they inherently operate through inference from ambiguous visual signals rather than direct visual processing.}
\label{tab:ocr}
\end{table*}

 \begin{figure}[ht]
    \centering
    \begin{subfigure}{0.45\textwidth}
        \includegraphics[width=\linewidth]{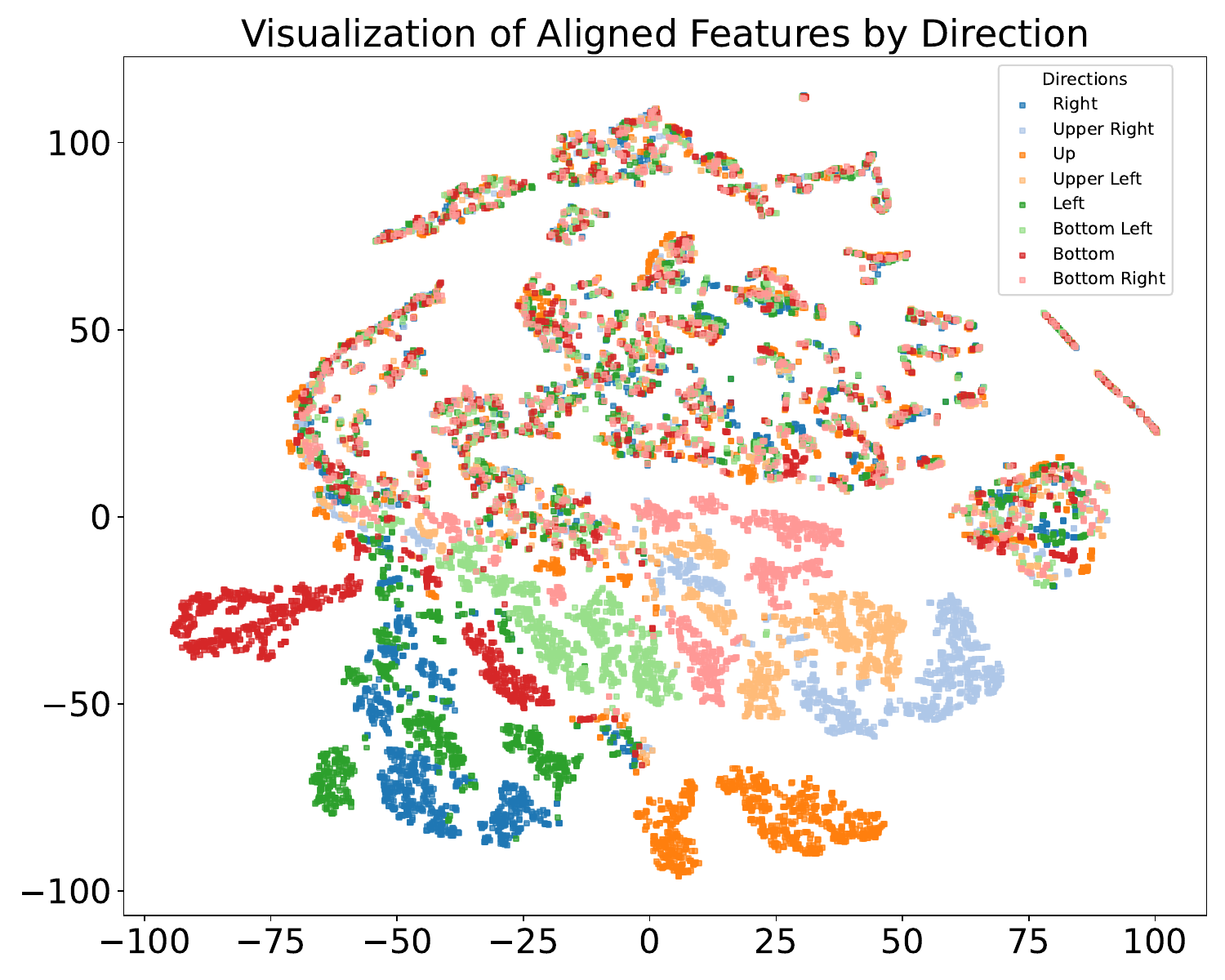}
        \label{fig:image1}
    \end{subfigure}
    \hfill
    \vspace{-1em}
    \begin{subfigure}{0.45\textwidth}
        \includegraphics[width=\linewidth]{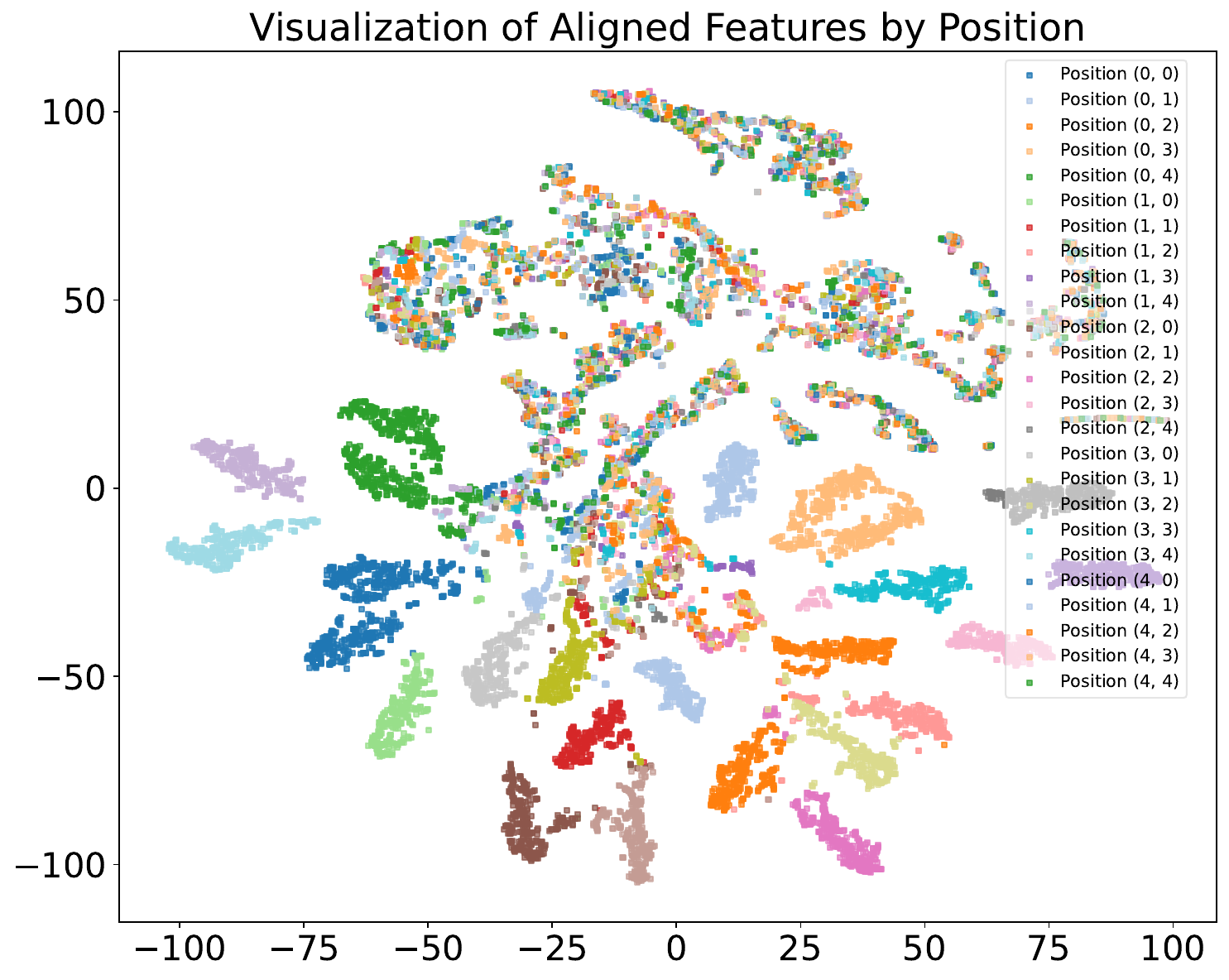}
        \label{fig:image2}
    \end{subfigure}
    \vspace{-1.2em}
    \caption{t-SNE visualization of aligned features under directional and positional variations, demonstrating partial sensitivity to visual variations in the feature space.}
    \label{fig:cluster_by_position_and_direction}
    \vspace{-0.2em}
\end{figure}

\section{Experimental Setup}
\label{sec:experiment-set-up}
We conduct comprehensive experiments across three distinct datasets to evaluate our model's performance under various conditions. The experimental settings for each dataset are detailed below.

\subsection{Coordinate Dataset}
For the coordinate dataset, we systematically vary four key parameters to thoroughly assess model performance. The point range parameter defines the spatial extent of the coordinate system, with values spanning from confined spaces ($[-5, 5]$) to broader ranges ($[-10, 10]$, $[0, 10]$, and $[0, 20]$). This variation allows us to evaluate the model's ability to handle different scales of spatial information.

To investigate the impact of visual aids on model performance, we experiment with both the presence and absence of reference lines and grid systems. These binary parameters (True/False) help us understand how additional visual context affects the model's coordinate understanding.

Regarding dimensionality, we focus our analysis on one- and two-dimensional coordinate systems. While three-dimensional coordinates were initially considered, they were ultimately excluded from our final experiments due to consistently poor model performance across preliminary tests.

\subsection{Path Dataset}
\label{sec:path-dataset}
In the path dataset experiments, we explore the model's capability to process and understand connected point sequences. We vary the complexity of paths by adjusting the number of points from 2 to 6, creating a progression from simple linear paths to more complex multi-point trajectories. The spatial distribution of these points is controlled through the same range parameters as the coordinate dataset: $[-5, 5]$, $[-10, 10]$, $[0, 10]$, and $[0, 20]$. For each unique combination of point count and range setting, we generate a substantial set of 100 images, ensuring robust evaluation across different configurations.

We evaluate the performance of path tracing and coordinate recognition through the following metric:

\paragraph{(1) Exact Match Accuracy (EMA)} A predicted path is considered correct only if it exactly matches the ground truth answer.
\paragraph{(2) Partial Match Order-Independent Accuracy (PM-IA)} In this metric, a point in the predicted path is deemed correct if its coordinates match any point in the ground-truth path. PM-IA is computed as the average accuracy across all positions.
\paragraph{(3) Partial Match Order-Sensitive Accuracy (PM-SA)} This metric marks a point in the predicted path as correct only if both its coordinates and position in the path match those of a point in the ground truth path. PM-SA is calculated as the average accuracy across all positions.
\paragraph{(4) Point Accuracy (PA)} This metric is the accuracy of the coordinate recognition.
\subsection{Object and Orientation Dataset}
The object dataset evaluates a model's understanding of object positioning and directionality. It comprises ten carefully selected object categories: eight animals (shiba dog, cat, bear, eagle, snake, panda, turtle, and fish) and two vehicles (car and plane). This diverse set enables assessment of model performance across varied object morphologies and complexities. In the linear probing experiment, the model performs a 90-class classification task based on these objects. 

In the orientation dataset, the model is instructed to detect the directions of the arrow in an image. There are in total 8 possible directions, including up, upper-right, right, bottom-right, bottom, bottom-left, left and upper-left. This comprehensive set of orientations ensures a robust assessment of the model's capacity for basic spatial reasoning and its interpretation of symbolic directional cues. In the linear probing setting, the model performs a 8-class classification task based on these objects.

Scale perception is tested through a comprehensive set of object-to-background ratios: 1/2, 1/3, 1/5, 1/10, 1/15, and 1/20. These ratios represent a wide spectrum from prominent objects (1/2) to more subtle presentations (1/20). Furthermore, we evaluate each object against two distinct background types: solid colors for controlled conditions and semantic images for real-world complexity.

The orientation aspect of our experiments encompasses eight distinct directions: the four cardinal directions (up, down, left, right) and their intermediates (top-left, bottom-left, top-right, bottom-right). This comprehensive directional coverage allows us to assess the model's ability to understand and interpret various object orientations.

Through these carefully designed experimental settings, we aim to provide a thorough and systematic evaluation of our model's capabilities across different aspects of visual understanding and spatial reasoning.

\subsection{Text Dataset}
\label{text-dataset}
We create char matrices of size 8*8, 16*16, 24*24, 32*32, 40*40 and 64*64. A target word is selected from one of the following, \texttt{[`dog', `cat', `bird', `lion', `tiger', `zebra', `monkey', `panda']}, and is positioned within the matrices. Except the object, the rest of the matrices are either asterisks (e.g. corresponding to the w/o BG setting) or random background words. We design 3 tasks to test the robustness of LLM in the cross-component analysis: (1) target word recognition, corresponding to object detection task in image; (2) coordinate recognition which corresponds to coordinate recognition; and (3) object counting, which is a fundamental skill needed in path tracing. By systematically varying the background contexts while maintaining the target object, we assess whether models truly recognize objects independently or merely rely on contextual associations. 

\section{Experimental Setup}
\label{app:exp}

\subsection{Evaluated Models}
\paragraph{Qwen-VL} The Qwen-VL model family represents Alibaba's cutting-edge vision-language model series. The family includes three main variants: the original Qwen-VL~\cite{Qwen-VL}, which established the foundation for vision-language processing; Qwen2-VL-7B, offering a balanced mid-size option; and Qwen2-VL-72B, the largest and most sophisticated version featuring state-of-the-art visual understanding capabilities and support for videos over 20 minutes long~\cite{wang2024qwen2vlenhancingvisionlanguagemodels}. The latest addition, QVQ-72B-Preview, serves as an experimental research model specifically focused on advancing visual reasoning capabilities~\cite{qvq-72b-preview}.
\paragraph{Molmo}
Molmo-7B-D~\cite{deitke2024molmopixmoopenweights} is an open-source vision-language model developed by the Allen Institute for AI, built on Qwen2-7B and utilizing OpenAI's CLIP as its vision backbone.
\paragraph{H2OVL}
The H2OVL-Mississippi~\cite{galib2024h2ovlmississippivisionlanguagemodels} model family are specifically designed for efficient on-device applications and privacy-focused use cases. The family consists of two specialized models: H2OVL-Mississippi-0.8B, a compact model optimized for text recognition that achieves state-of-the-art performance on OCRBench, and H2OVL-Mississippi-2B, a model for general vision-language tasks including image captioning and visual question answering.
\paragraph{Phi-3} 
Microsoft's Phi-3-vision and Phi-3.5-vision~\cite{abdin2024phi3technicalreporthighly} represent a significant advancement in multimodal AI. Phi-3.5-vision, the latest iteration, is a lightweight yet powerful model featuring a 128K token context length and support for both single and multi-image processing.

\vspace{-0.3em}
\paragraph{InternVL}
The InternVL family includes: Mono-InternVL~\cite{luo2024monointernvlpushingboundariesmonolithic}, which established the foundation with its vision-language capabilities; InternVL-2~\cite{chen2024internvl}, which expanded the model sizes and improved performance; and InternVL-2.5~\cite{chen2025expandingperformanceboundariesopensource}, the latest iteration that introduces significant architectural and training improvements.
\vspace{-0.4em}
\paragraph{LLaVA}
LLaVA-1.5~\cite{liu2023visualinstructiontuning} uses pretrained CLIP~\cite{radford2021learningtransferablevisualmodels} and Vicuna language model as the backbone, establishing the foundation with impressive performance across 12 benchmark datasets. LLaVA-OneVision-Qwen2~\cite{li2024llavaonevisioneasyvisualtask} pushes performance boundaries across single-image, multi-image, and video scenarios while enabling strong task transfer capabilities. LLaVA-1.6\cite{liu2024llavanext}, or LLaVA-NeXT, further enhances capabilities with increased input resolution, improved visual reasoning, and enhanced OCR capabilities.
\paragraph{LLaMA}
LLaMA 3.2 Vision models ~\cite{dubey2024llama3herdmodels} represent Meta's latest advancement in multimodal AI, introducing vision capabilities to the LLaMA family for the first time. The 11B and 90B parameter versions are specifically designed to handle both text and image inputs, featuring a novel architecture that integrates image encoder representations into the language model.

\paragraph{GLM-4}
GLM-4V-9B~\cite{glm2024chatglm} supports high-resolution image processing at 1120*1120 pixels and enables dialogue capabilities in both Chinese and English.
\paragraph{MiniCPM-V}

MiniCPM-V~\cite{yao2024minicpmvgpt4vlevelmllm} is a series of multimodal large language models (MLLMs) designed specifically for deployment on end-side devices like mobile phones and personal computers.
\subsection{Implementation Details}
Our experiments are conducted on 8 NVIDIA H100 GPUs. All models maintain their original parameter configurations during inference, with an average processing speed of 525 tokens per second.

To examine spatial reasoning capabilities analogous to those in LVLMs, we design a suite of tasks that evaluate text-based object recognition, counting, and spatial analysis. Using a matrix-structured text input that simulates idealized visual encoding, we assess three fundamental capabilities: (1) target token identification amid background elements, (2) frequency quantification of target tokens, and (3) spatial localization of target tokens within the matrix. The dataset specifications and distributions are detailed in Appendix~\ref{text-dataset}. 

\section{Visually Conditioned Soft Prompt}
\label{app:soft_prompt}
In this section, we explain why aligning visual features to the language model's embedding space through multimodal alignment module can be viewed as a form of visually conditioned soft prompt.
Soft prompting prepends learned vectors to the language model's input, optimizing these vectors during training to achieve desired tasks. These vector prefixes learn latent instructions during the tuning process:
\vspace{-0.2em}
\begin{equation}
\begin{aligned}
\mathbf{h}_t &= LM([\underbrace{P}_{\text{prompt}} ; \mathbf{x}]_{<t}), \\
&\text{where } P \in \mathbb{R}^{l \times d}, \quad \mathcal{L} = \sum_t -\log p(x_t|\mathbf{h}_t)
\end{aligned}
\end{equation}
where $P$ represents the learned soft prompt vectors and $\mathbf{x}$ is the input token embeddings.
In LVLM's multimodal alignment process, extracted visual features are mapped to the language model's embedding space through a neural network based multimodal alignment module. The module is trained to transform visual features into embedding space that serve as special task prefixes, guiding model responses to image-related queries::
\begin{equation}
\begin{aligned}
\mathbf{h}_t &= LM([\underbrace{f_{\theta}(E_{v}(I))}_{\text{visual prompt}} ; \mathbf{x}]_{<t}), \\
&\text{where } \mathcal{L} = \sum_t -\log p(x_t|\mathbf{h}_t, I)
\end{aligned}
\end{equation}
where $f_{\theta}$ is the alignment network and $\mathbf{v}$ represents the visual features.
Fundamentally, both approaches optimize vector prefixes independent of the discrete language embedding matrix. The key difference lies in their optimization objectives: soft prompts aim to generate task-relevant responses based on the dataset, learning fixed prefixes that remain constant after training. In contrast, while aligned visual features are also trained on the entire dataset, they produce dynamic, image-specific prefixes during inference. Therefore, aligned visual features can be considered a special case of soft prompts - visually conditioned soft prompts - sharing the characteristic of residing in a representation space distinct from the language embedding space.

\section{Failure Examples}
\label{sec:examples}
\begin{itemize}[leftmargin=*, itemsep=2pt]
    \item \textbf{ChatGPT:} Sorry, I don’t see the arrow in the image.

    \item \textbf{Claude:} A sleek fish glides through the harbor waters, its scales catching glints of sunlight filtering through the surface. The backdrop features wooden pier posts rising from the water and moored fishing boats gently bobbing in the distance. The water has a deep blue-green tint typical of a working harbor, with small ripples creating patterns of light and shadow. The fish appears to be navigating between the shadows cast by the docks, perhaps searching for food or shelter among the port's underwater structures.

    \item \textbf{Gemini:} 
    \textbf{Reasons:}
    \begin{itemize}
        \item \textbf{Image Information:} The image only shows an animal on the water's surface, without indicating the specific direction of the animal's head, body, or tail.
        \item \textbf{Lack of Reference Points:} There are no other objects or background elements to help determine the animal's orientation.
    \end{itemize}
    Therefore, it is impossible to determine the animal's orientation from the information provided in the image.

    If you can provide more information about the image, such as the shooting angle or the animal's posture, it may be possible to determine its orientation more accurately.

    \item \textbf{InternVL:} This image shows a pure white background without any recognizable objects or text. It is a simple plane with no complex designs or patterns.
\end{itemize}

\end{document}